# MMFL-Net: Multi-scale and Multi-granularity Feature Learning for Cross-domain Fashion Retrieval

Chen Bao[1], Xudong Zhang[2], Jiazhou Chen[1], Yongwei Miao[3]

1. College of Computer Science and Technology, Zhejiang University of Technology, Hangzhou, China
2. School of Information Science and Technology, Zhejiang Shuren University, Hangzhou, China
3. School of Information Science and Technology, Hangzhou Normal University, Hangzhou, China

(ywmiao@hznu.edu.cn; xdzhang@zjsru.edu.cn; )

## Abstract

Instance-level image retrieval in fashion is a challenging issue owing to its increasing importance in real-scenario visual fashion search. Cross-domain fashion retrieval aims to match the unconstrained customer images as queries for photographs provided by retailers; however, it is a difficult task due to a wide range of consumer-to-shop (C2S) domain discrepancies and also considering that clothing image is vulnerable to various non-rigid deformations. To this end, we propose a novel multi-scale and multi-granularity feature learning network (MMFL-Net), which can jointly learn global-local aggregation feature representations of clothing images in a unified framework, aiming to train a cross-domain model for C2S fashion visual similarity. First, a new semantic-spatial feature fusion part is designed to bridge the semantic-spatial gap by applying top-down and bottom-up bidirectional multi-scale feature fusion. Next, a multi-branch deep network architecture is introduced to capture global salient, part-informed, and local detailed information, and extracting robust and discrimination feature embedding by integrating the similarity learning of coarse-to-fine embedding with the multiple granularities. Finally, the improved trihard loss, center loss, and multi-task classification loss are adopted for our MMFL-Net, which can jointly optimize intra-class and inter-class distance and thus explicitly improve intra-class compactness and inter-class discriminability between its visual representations for feature learning. Furthermore, our proposed model also combines the multi-task attribute recognition and classification module with multi-label semantic attributes and product ID labels. Experimental results demonstrate that our proposed MMFL-Net achieves significant improvement over the state-of-the-art methods on the two datasets, DeepFashion-C2S and Street2Shop. Specifically, our approach exceeds the current best method by a large margin of +4.2% and +11.4% for mAP and Acc@1, respectively, on the most challenging dataset DeepFashion-C2S.

**Keywords**   Visual search, Multi-scale features, Multi-granularity features, Content-based image retrieval(CBIR), Feature Learning

## 1. Introduction

With the rapid growth of fashion e-commerce and fast-fashion community, the online fashion shopping has come into an exponential rise. Compared with brick-and-mortar stores, an increasing number of customers have become a part-and-parcel of online shopping[1]. Additionally, the e-commerce market has excellent commercial and cultural values. Building a user-centric intelligent fashion search system to quickly accommodate customers and accurately retrieve large-scale databases is the key to fostering the core competitiveness of large fashion e-commerce websites. Large e-commerce websites, such as Ebay.com [2], have developed a content-based image retrieval e-commerce platform to facilitate online clothing shopping.

   Content-based clothing image retrieval[1, 3] (e.g., visual similarity or instance-level image retrieval) has a wide range of applications, such as visual recommendation or visual search; it has emerged as an active research topic in both academia and the fashion industry. Cross-domain clothing image retrieval between customers and online shopping stores (C2S), which is closely related to visual search, is considered a subtask of instance-level image retrieval. It has potential value in a variety of fashion-related real-world applications, such as providing a street snapshot or Instagram post depicting a clothing item as a query and matching the vendor's catalogue (gallery) with visually similar or identical garment items to recommend related products. Other applications include: fabric search and retrieval in fashion industry scenarios, which is applied to fabric purchase and garment design, and offline audit analysis in the new apparel retail surveillance scenario to intelligently classify and identify clothing products. In addition, modeling cross-domain clothing retrieval faces two main challenges: (1) It must identify heterogeneous domain situations, that are specific to the separate domains of the consumer and shop. There is a significant domain discrepancy between the customer's unrestricted realistic garment image and the retrieved image from shop's catalogue, which could be caused by viewpoint changes, poor lighting, background clutter, variations in pose and occlusion, etc. (2) Clothing images always contain many different attributes. For example, clothing includes global appearances (e.g., color, style, category) and also detailed attributes (e.g., texture, material, pattern, collar, hem). Simultaneously, the drape of clothing means that it has high deformability and irregular appearance. Thus, the clothing images always exhibit high intra-class variance and inter-class similarity, making the open set task of C2S clothing retrieval extremely challenging.

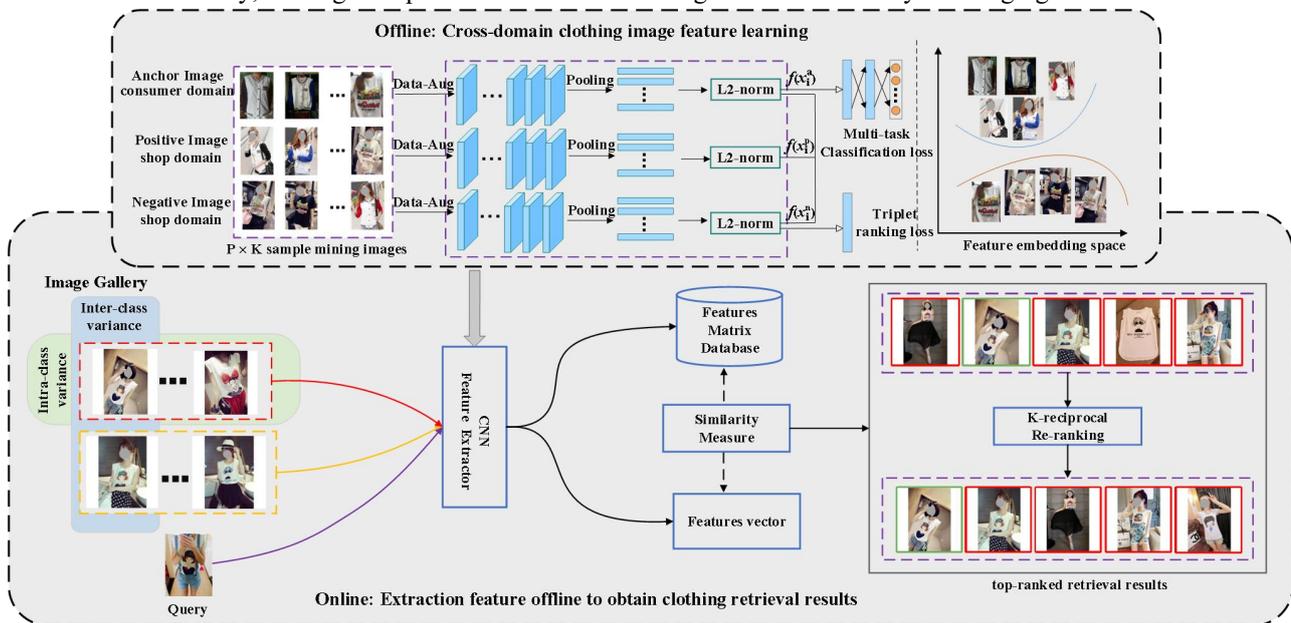

**Fig. 1** Schematic diagram of clothing image retrieval

Conventional approaches for clothing image retrieval always rely on the selection of hand-crafted features, whilst extracting the visual properties of shallow layers, and preserving the semantics of visual phrases using machine learning models, such as SVM[4]. However, these shallow-layer visual features lack the semantic information, visual appearance, and local descriptors of actual clothing images. The extracted feature descriptors always lack a specific learning ability, and its generalization ability is insufficient. However, the surge in research interest in the deep convolutional neural networks (CNNs), and development of fashion benchmarks, such as ACWS [5], Street2Shop [6], DARN [7], DeepFashion [8], and FindFashion [9], have boosted the C2S clothing image retrieval. The CNN-based image retrieval can automatically learn the complex cross-domain retrieval scenarios by creating an image-feature indexing database, calculating the similarity between continuous vectors in the corresponding feature embedding spaces, using the cosine distance or Euclidean distance to measure their similarity of fashion items, and finally returning to a visual search for results retrieval. Fig. 1 shows the schematic diagram of clothing image retrieval. At present, CNN-based clothing image retrieval has greatly surpassed the performance of traditional methods, which leads to its popularity in fashion-related research, especially for fashion attribute classification and recognition [5, 10], clothing image retrieval[8, 9, 11], and fashion compatibility prediction [12, 13], etc.

The majority of CNN-based methods can effectively compute the overall similarity in a general embedding space, that is, they can map a holistic representation and extract global salient features. However, it is difficult to learn subtle visual similarities, detailed style features, or discriminative details of local regions. Employing a global similarity pipeline to adapt the common inter-class properties or large intra-class variance could lead to the sub-optimal retrieval performance of image retrieval. In fact, the previous CNN-based methods based on local feature vectors always applied the predefined semantic information to identify local regions and learn local information representation[14]; however, it is difficult to learn clothing images with their large variances. Moreover, for the task of cross-domain C2S clothing image retrieval tasks, the local features can highlight identity-related detail information, fuse the global and local aggregation information, enhance clothing image instance discrimination, and improve the distribution of clothing image features by providing a complementary mechanism to describe clothing images. In addition, owing to the pyramid like hierarchical structure of CNNs, we can extract both low-level and high-level features of underlying clothing images, wherein the low-level features contain rich spatial structure information but lack semantic information. In contrast, the high-level features have abundant semantic information, however, they are less detailed information which lead to a semantic- spatial gap. Therefore, a goal of field-relevant research is to effectively fuse both high-level and low-level features for extracting the robust and discriminative visual features of clothing images.

In this paper, we take a feature representation approach and propose a multi-scale and multi-granularity feature learning network (MMFL-Net) to jointly learn global-local aggregation feature representations by combining the global and local view information. First, we design a semantic-spatial feature fusion part (SFFP) comprising a pyramid-like semantic modulation block (PSMB) and context-aware feature attention extraction (CFAE) modules. We use PSMB to apply top-down and bottom-up bidirectional multi-scale feature fusion approach to effectively fuse the high-level semantic features with low-level spatial structure features to bridge the semantic-spatial gap. We further apply the CFAE module to strengthen the multi-scale features with the global context information of the clothing images. Next, we propose a multi-branch deep network model to capture the global salient, part-informed, and local detailed information by integrating the similarity learning of coarse-to-fine embedding with a multi-granularity network. Finally, we optimize an improved trihard loss, center loss, and multi-task classification loss to improve the discriminative ability of the cross-domain retrieval model. The experimental results demonstrate that our proposed MMFL-Net exhibits competitive performance improvement over the state-of-the-art methods on the two datasets, DeepFashion-C2S and Street2Shop.

The main contributions of our proposed approach are summarized as follows:

- **A novel feature learning network.** We propose a novel network model—MMFL-Net to learn global-local aggregation feature representations of clothing images, which can effectively extract the global salient, part-informed, and local detailed information by combining the similarity learning of coarse-to-fine embedding with a multi-granularity network. The MMFL-Net model outperforms most state-of-the-art methods and achieves superior performance in cross-domain C2S fashion retrieval.
- **Effective multi-scale feature fusion structure.** We design a SFFP that comprises PSMB and CFAE modules, which apply top-down and bottom-up bidirectional multi-scale feature fusion strategy to bridge the semantic-spatial gap. The SFFP is verified to be capable of effectively fusing semantic-spatial information, generating multi-scale visual feature representations with a global context information.
- **Improved hard mining trihard loss.** We present an improved trihard loss in conjunction with a multi-task classification loss that can jointly optimize the representation learning of embedding space interfaces and the deep metric learning of embedding space clustering. This is done with respect to the specific attributes and PID labels, thereby improving the intra-class compactness and inter-class discriminability of clothing image features.

The rest of this paper is organized as follows. Section 2 gives a brief review of the related work. The proposed approach is shown in detail in Section 3. Section 4 validates the implementation of the experiment and evaluation results. Finally, the paper is concluded the work and proposed future research in Section 5.

## 2. Related work

### 2.1 Clothing images feature learning

**Conventional methods.** Cross-domain clothing retrieval, which aims to compute the similarity between fashion items, is currently popular with e-commerce and online fashion shopping researchers. Early explorations of cross-domain retrieval relied on hand-crafted features[1], such as global feature-based descriptors (e.g., color, shape, and texture), local feature-based descriptors (e.g., SIFT, HoG, and LBP), bag-of-visual-words (BoVW), and VLAD or FV aggregated local patches to construct image feature representations. Liu et al. [15] first addressed the problem of cross-scenario clothing retrieval and obtained a human body part region by using a human body pose estimator. They also utilized feature fusion (e.g., HoG and LBP) to extract part-based clothing image representations and an unsupervised similarity transfer matrix to transfer the representation of street snapshot images to the representation space of clothing images for online shopping. In the 3D shape retrieval task[16, 17], which corresponds to fashion retrieval task, texture is an essential feature of 3D object appearance. Biasotti et al. [16]introduced a comparative study of six methods for the classification of texture 3D mesh models. However, the methods based on hand-crafted features and semantic-preserving visual phrases are limited by the expressive ability of these features.

**Global feature learning.** CNNs have proven to be powerful feature embeddings, which has led to significant progress with C2S clothing retrieval. Kiapour et al. [6] proposed a WTBI street-to-shop clothing image retrieval model which adopted the pre-trained AlexNet to output fully-connected layer as the global feature representation of clothing image. Huang et al. [7] presented a DARN architecture that employs semantic attribute information by stacking NIN convolutional units to learn powerful visual representations. They applied the triplet loss to measure the feature similarity of street snapshots and online shopping clothing images. Kuo et al. [18] utilized global features and NetVLAD, combined with rank-based candidate selection, for consumer-to-shop clothing retrieval. Kucer and Murray [19] proposed a multi-domain fashion item retrieval model that uses a classic object detection model to detect fashion items. They further designed a three-stream Siamese architecture in an ensemble model comprising two single models, one of which was trained using triplet loss and AP loss. Wieczorek et al. [11] introduced a strong baseline for fashion retrieval capability using a person Re-ID model, which uses CNN-extracted features as global feature vectors. For these existing methods, clothing images are always encoded as coarse-grained global feature vectors, and their global representations can be adopted to perform clothing retrieval, ignoring the expression of local information, which lead to sub-optimal retrieval performance.

**Local feature learning.** Apart from global feature representations, many existing approaches focus on local feature learning to extract fine-grained local features for fashion retrieval[8, 20, 21]. Several representative local-based approaches have been introduced, such as semantic component and part-informed approaches. The semantic component approach utilizes a clothing landmark estimation model or a human parsing model for local feature localization on the relevance of the salient sub-regions of clothing images. Liu et al. [8] proposed a FashionNet that jointly optimizes the objectives for attribute prediction and clothing recognition or retrieval. FashionNet can fuse the local features of landmark alignment with the global semantic features of a clothing image, and also learn both global feature representation and local semantic information. Lang et al. [20] proposed a PS-Net for plagiarized clothing retrieval that utilized landmarks to guide the learning of local regional feature representations and compared fashion items regionally using triplet-based loss. Liu et al. [21] utilized semantic labels located in fine-grained human parsing and generated semantic part-based stronger feature representations by fusing the semantic masks with the generated features of convolutional layers. However, owing to its disparities in the part-based estimation of local feature on semantic components, it is necessary to delineate the component region exactly, which is essential for model generalization and could achieve desirable retrieval results.

Part-informed approaches extract deep feature maps from a CNN using a partition strategy, and each part-based approach represents a clothing image component. These approaches were proposed using the part-based convolutional baseline (PCB) [22] partition model and a multiple granularity network (MGN) [23] for Re-ID. Consequently, this idea is directly portable to cross-domain retrieval for obtaining discriminative feature representations. Kuang et al. [9] proposed a GRNet to develop a similarity pyramid that learned the similarity using both global and local representations at multiple scales. They also propagated and updated global and local similarities via graph reasoning. Luo et al. [24] proposed integrating global and local information in a multiple granularity network consisting of three branches with spatial constraints between different parts. Although this approach achieved better retrieval results on canonical datasets, the misalignment of local feature remains complicated.

**Attribute-guided feature learning.** Attribute-based approaches assisted the models in finding intra-class and inter-class correlations. Fashion attributes are adopted to define mid-level semantic visual concepts and train semantic attribute labels for the generalization and discriminative ability of different fashion objects. Chen et al. [25] considered the differences of the semantic attribute subcategories and further decomposed the attributes of fashion items into detailed attributes (e.g., sleeve, color, collars, and fabric). Ak et al. [26] presented a learning framework of general attribute representations by exploiting attribute activation maps generated by a part-based triplet ranking loss. Hou et al. [27]proposed a combination of multi-label attribute classification and visual-semantic consistency loss to obtain disentangled feature representations for attribute manipulation interactive fashion retrieval. Although clothing attribute recognition is not the primary target of our work, our clothing retrieval model is closely related to attributes because the primary objective involves learning the local detailed feature representations with attributes.

**Attention-based feature learning.** The attention mechanism can guide a network to focus on the region of interest features and thereby ignore the irrelevant features. The mainstream attention models have been divided into three types: spatial attention, channel attention, and self-attention. Spatial attention models the spatial domain information of the feature maps corresponding to the spatial transformation. Ji et al. [28] presented a CtxYNet network for cross-domain clothing retrieval wherein the network introduced a spatial attention model based on attribute guidance to facilitate the extraction of clothing images representations. Luo et al. [29] proposed a DMCH network, which jointly embeds sequential attribute learning and spatial attention-aware feature representation for cross-domain retrieval. However, spatial attention mechanisms usually focus on the most prominent areas of an image, which can lead to ignore some useful local detail features. Channel attention models the correlation between different channels (feature maps) by enhancing channel-wise convolutional activations. Wang et al. [30] proposed a VAM model that applies the low-level features of the network branch learned by the channel attention mechanism to obtain more discriminative global feature descriptors. Dong et al. [31] proposed a novel approach using attribute-aware spatial and channel attention, which can learn features of the corresponding fine-grained image attributes thus to improve the similarity discrimination of clothing images. Self-attention is a variant of the attention mechanism that efficiently captures long-range dependencies by modeling the global context through a self-attention mechanism. Lu et al. [13] inserted of a non-local (NL) block into the ResNet-50 architecture to capture the correlation between each location on a feature map and equip the network with globally refined features. Tan et al. [32]adopted RRTs transformer for local and global feature extraction and reordered the matched images using a supervised approach, thereby improving the recognition accuracy.

## 2.2 Multi-scale feature fusion

Most deep CNNs were adopted to represent the multi-scale fusion features for obtaining more discriminative features. Newell et al. [33] proposed an hourglass network that employs continuously stacking the bottom-up and top-down units to improve features at different scales, from local to global. Lin et al. [34] proposed a feature pyramid network (FPN) that adopts a top-down pathway and lateral connections for constructing multi-scale pyramid features. Following this concept, Liu et al. [35] proposed PANet, which strengthens overall features using additional bottom-up pathway localization information. EfficientDet [36] is a highly effective, weighted bidirectional FPN (BiFPN) that applied learnable weights (similar to attention) to improve the balance of feature information corresponding to different scales and repeatedly perform bidirectional multi-scale feature fusion. In MLFPN [37], Zhao et al. used TUM, FFM and SFAM modules to extract multi-scale features and fuses multi-level features. In addition, Zhang et al. [38] provided the SEB module, which multiplies the element-wise product at deep and low levels at the corresponding positions. In this paper, we design a semantic-spatial feature fusion part (SFFP) for bidirectional multi-scale feature fusion.

## 3. Proposed network

### 3.1 Overview

Our work aims to train an end-to-end cross-domain visual similarity model for C2S clothing retrieval. A working example of our model is consumer-to-shop retrieval that uses an unrestricted user photo to match the similarity in the semantics, such as sleeve length and collar, and appearance, such as color and texture, of a garment item in online clothing shops to bridge the gap between the shop and consumer domain. Specifically, given an image of clothes $I$ and image-level multi-label annotated data. We propose to learn feature representations $f_I = \mathcal{F}(I;\theta)$. Here, $\mathcal{F}(\cdot)$ is a parametric mapping function that computes the image representation, and $\theta$ represents the trainable parameters of $\mathcal{F}(\cdot)$. Thus, there are two sub-tasks, namely, designing a CNN-based feature embedding network and the similarity metric leaning.

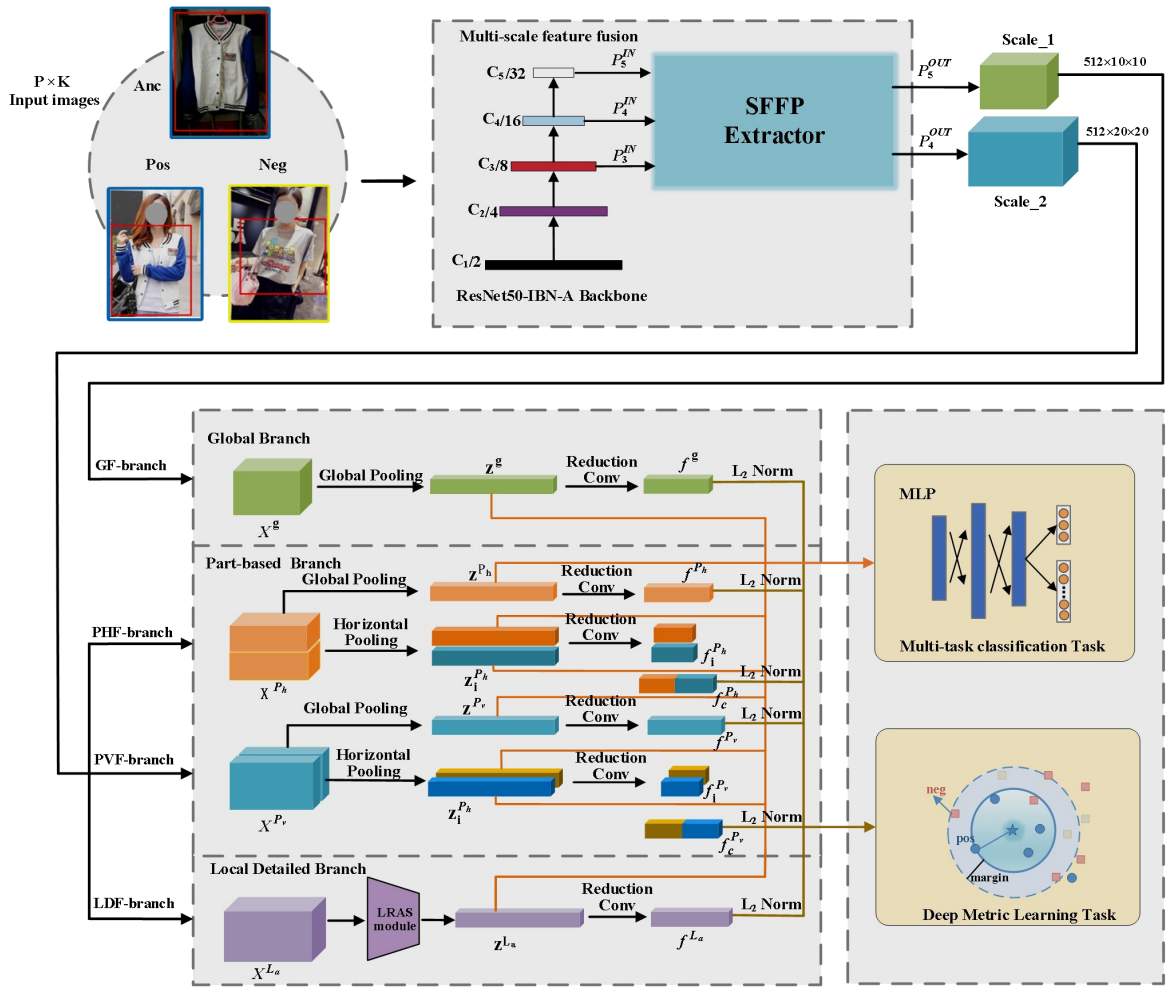

**Fig. 2** The overall architecture of our MMFL-Net, which consists of four branches, that is, global feature branch (GF-branch), part-based horizontal-partition feature branch (PHF-branch), part-base vertical-partition feature branch (PVF-branch), local detailed feature branch (LDF-branch). GF-branch extract global salient information; PHF-branch or PVF-branch are part-based structures to capture local part-based information; LDF-branch can generate local detailed features. During the training phase, all feature extractor branches are combined with multi-task classification loss and deep metric loss. During the inference phase, fully-connected layer on the PIDs task is concatenating as a final feature embedding.

The overall architecture of our MMFL-Net is shown in Fig. 2. Our approach adopts mini-batch images to train the MMFL-Net. Accordingly, we select ResNet50-IBN-A as our backbone network. In each mini-batch, clothing images are random horizontal flipping, rotating, color jittering, padding resize, and warped to 320×320 pixels as input images.

Hence, 512-dim deep feature response maps can be extracted by the SFFP extractor, which apply top-down and bottom-up bidirectional multi-scale feature fusion strategy to bridge the semantic-spatial gap. Then, the multi-scale feature response maps with the global context information are partitioned and pooled to obtain multi-granularity features that pass through a four-branch network comprising a global feature branch (GF-branch), a part-based horizontal partition local feature branch (PHF-branch), a part-based vertical partition local feature branch (PVF-branch), and a local detailed feature branch (LDF-branch). We construct the GF-branch, PHF-branch, and PVF-branch to maintain the coarse-scale information and part-scale abstract component information of clothing images, respectively. The LDF-branch adopts a channel-wise attention to generate the fine-grained and semantic correlation features. In order to obtain discriminative features, clothing attribute prediction, PID classification and similarity metric learning training are simultaneously performed in each-branch of MMFL-Net and the multi-task feature learning is jointly optimized. Among them, the Euclidean distance between attribute prediction and optimization features is added to suppress semantic drift and improve the presentation of different clothing images. The improved trihard loss is better at minimizing the optimal metric of intra-class and inter-class distance. Hence, we can jointly train and optimize all four branches with network extract global-local aggregation feature embedding to perform better generalizations. In the inference phase, the task for the PID in the first fully-connected layer, extracts the feature similarity score of clothing images, computed in terms of cosine similarity, as final feature embedding. Meanwhile, the re-ranking post-processing method is combined to obtain more accurate retrieval results.

### 3.2 Feature learning network (MMFL-Net)

#### 3.2.1 Global feature branch (GF-branch)

Inspired by the recent work [11], we use a GF-branch to extract global salient information. The GF-branch consists of three steps. First, the SFFP extractor convolutional activation tensor $X^g \in \mathbb{R}^{512\times10\times10}$, global average pooling (GAP), and global max pooling (GMP) are applied to construct a global feature vector $\mathbf{z}^g \in \mathbb{R}^{512\times1\times1}$, to capture global information and compute the multi-task classification loss. Then, using the reduction convolutional layer, the global feature vectors are normalized with the BatchNorm (BN) layer and the ReLU activation function to obtain a 256-dim feature embedding $f^g$. Finally, $f^g$ is used during training as input to the trihard and center losses. Global feature vector $\mathbf{z}^g$ is applied for attribute prediction and PID classification. During the inference phase, the first fully-connected layer through BN layer on the PID task of MLP hidden layer is used for clothing retrieval, and the cosine similarity distance is selected to calculate similar distances.

#### 3.2.2 Part-based feature branch

Local information has recently played an increasingly important role in fashion retrieval tasks. To learn discriminative part-informed features, we design a local feature extraction branch based on abstract component features similar to the PCB and MGN, which can implement horizontal and vertical partitions on feature maps. To this end, we use 2D adaptive average pooling and adaptive max-pooling operations, and we spatially down-sample the convolutional activation tensor $X^{P_h} \in \mathbb{R}^{512\times20\times20}$, from the SFFP feature extractor into a column feature vector $\mathbf{z}^{P_h} \in \mathbb{R}^{512\times2\times1}$. Then, we slice the horizontal partition of the convolutional feature maps $X^{P_h}$, which has a global receptive field with rich contextual information for each feature partition $\mathbf{z}_i^{P_h} \in \mathbb{R}^{512\times1\times1} (i=1,2)$. We use the reduction convolutional layer with BN and ReLU to reduce the dimension, that is, from 512-dim $\mathbf{z}_i^{P_h}$ to 128-dim $f_i^{P_h}$, which can suppress the overfitting. We also concatenate $f_i^{P_h}(i=1,2)$ vectors to obtain the 256-dim feature vector $f_c^{P_h}$, thus for trihard loss training.

For horizontal partition, the model localizes the local region (e.g., neckline and hem). Similarly, we exploit the vertical partition strategy to take the input convolutional feature maps $X^{P_v}$, use the 2d adaptive average pooling and 2d adaptive max pooling operations, and obtain a row feature tensor $\mathbf{z}^{P_v} \in \mathbb{R}^{512\times1\times2}$. Then, we slice $\mathbf{z}^{P_v}$ vertically by column, each generating a 512-dim feature vector $\mathbf{z}_i^{P_v} \in \mathbb{R}^{512\times1\times1} (i=1,2)$. We reduce the dimensions, resulting in the 128-dim feature vector $f_i^{P_v}$. For vertical partitioning, the model can localize the sleeve-length local sub-region. We also concatenate $f_i^{P_v}$ ($i=1,2$) vectors to gain the 256-dim vector $f_c^{P_v}$, for trihard loss training. Meanwhile, considering that using the local features directly may cause the PHF-branch or PVF-branch to ignore the overall information, we retain the global description features $\mathbf{z}^{P_h}$ and $\mathbf{z}^{P_v}$, to train as input to JARN, respectively.

### 3.2.3 Local detailed feature branch (LDF-branch)

To fully exploit the important channel-wise information in the convolutional feature maps, we use the LDF-branch, which extracts local detailed features based on channel-wise attention. First, we utilize the ECA [39] module to build the local attention re-weighting module, and then use the local residual attention selection (LRAS) module to incorporate a local attention selection mechanism which obtains the top-K key part activation feature maps for generating the discriminative local feature representations.

(1) *Local attention re-weighting module*. Instead of the SE block, we adopt the ECA module to avoid the necessity of dimensionality reduction while learning channel-wise attention. An appropriate cross-channel interaction can maintain the performance and significantly reduce the complexity of the model. Let $X^{L_a} \in \mathbb{R}^{512 \times 20 \times 20}$ be the feature response maps, which contain the distributed representation channel feature maps $X^{L_a} = \{X_i^{L_a}\}$ ($i=1,2,...,C$). The symbol $C$ indicates the number of channels. Considering each channel and its k-neighbors in the attention prediction of the channel, we model the channel-wise attention by re-weighting the channel features using the inter-channel feature relations. Following the ECA-Net, an ECA module comprises two steps: a fast 1D convolution and an adaptively selected of kernel size $k$. Hence, the re-weighting channel feature vector by efficient channel attention is formulated as follows,

$$M^c = \sigma(C1D_k(\text{GAP}(X^{L_a}))) \tag{1}$$

where $C1D_k$ means the 1D convolution, $\sigma$ is the sigmoid function, $k$ denotes the adaptive selection of kernel size, and $M^c$ is equivalent to a channel attention. Accordingly, the activation channel masks and feature maps $X^{L_a}$, generate re-weight adjusted attention features using an element-wise multiplication.

(2) *Local residual attention selection module*. The idea of attention residual learning was presented in the literature [40], where the feature response maps of different layers in a deep network are different. Low-level features concentrate on the background and other regions, whereas high-level features locate the salient regions of the image. Like the idea of residual learning, by stacking the structure of residual attention module, the attention soft mask is constructed as an identity mapping which can be used not only for feature selection in a feed-forward structure, but also for gradient weight updating during back-propagation. Inspired by Wang et al. [40], we provide a LARS module with two streams (i.e., trunk stream and attention mask stream), as shown in Fig. 3. The trunk stream corresponds with its mask branch to learn the attention of its local residual features. First, we generate 256-dim features using a 1×1 convolutional layer and then accumulate the pixel response intensity on the attention feature maps in the channel dimension to obtain the score $\bar{m}_c$, and select the activation map. We sort the intensity values of all channels by their pixel responses. According to the top-K strategy, we select a high score $\bar{m}_c$, as the key-part activation map to achieve unsupervised localization of the semantic part-level information. Thereafter, we can improve the output of the local residual attention module $F$ as follows,

$$F_i\big|_{i=1}^{k} = (1 + M_i(X^{L_a})) * T_i(X^{L_a}) \tag{2}$$

where $F_i\big|_{i=1}^{k}$ represents the k-*th* key-part activation map, $T_i(X^{L_a})$ denotes the identity mapping of the trunk stream residual unit, and $M_i(X^{L_a})$ denotes the attention mask stream soft mask. For the key-part activation map $F_i \in \mathbb{R}^{H \times W \times C}$, we use GAP to generate the quantization feature vector $f_i^{L_a} \in \mathbb{R}^{C \times 1 \times 1}$, and concatenate all $k$ feature vectors by their per-channel dimensions to model the activation feature map $f^{L_a} \in \mathbb{R}^{C \times k \times 1}$.

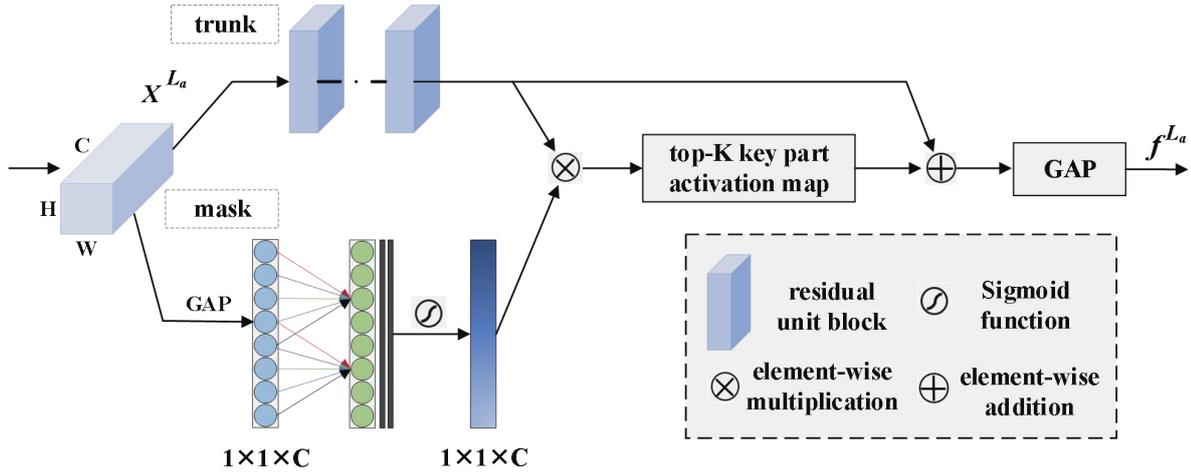

**Fig. 3** The local residual attention selection (LRAS) module

### 3.3 Semantic-spatial feature fusion part (SFFP)

#### 3.3.1 Pyramid-like semantic modulation block

In the pyramid-like CNN hierarchical structure, high-level features have more semantic contextual information, whereas low-level features have more spatial appearance information. We follow the later fusion strategy to strengthen feature propagation and encourage feature reuse to bridge the semantic-spatial gap. Hence, we leverage a pyramid-like semantic modulation block (PSMB) to combine more contextual information and multi-scale feature representations. As shown in Fig. 4, our PSMB consists of two sub-structures: (1) a BiFPN pyramid-like multi-scale feature fusion and (2) a resolution-aware fusion block. BiFPN pyramid-like multi-scale feature fusion applies their units by stacking the architecture, wherein our pyramid-like structure is a ResNet50-IBN-A backbone network. This ResNet50-IBN-A architecture can extract four stages of feature response maps $C_2$, $C_3$, $C_4$, and $C_5$, via a bottom-up side pathway. Three feature response maps $C_3$, $C_4$, and $C_5$, are chosen for feature fusion. First, we use a $1\times1$ convolution to reduce the channel dimension to 256-dim, leading $P_3^{IN}$, $P_4^{IN}$, and $P_5^{IN}$ through their lateral connections have resolution of 1/8, 1/16, 1/32, respectively, compared with the size of the input images. BiFPN pyramid-like structure adopts bidirectional multi-scale connections and implements combination of multi-scale weighted method and feature fusion at different resolutions. BiFPN takes level-3 to level-5 input features $P^{IN}=(P_3^{IN},P_4^{IN},P_5^{IN})$, as the input and then aggregates the multi-scale features in a top-down and bottom-up manner as follows,

$$P_4^{TD} = Conv\left(swish\left(\frac{w_1 \cdot P_4^{IN} + w_2 \cdot \text{Upsample}(P_5^{IN})}{w_1 + w_2 + \varepsilon}\right)\right) \quad (3)$$

$$P_4^{OUT} = Conv\left(swish\left(\frac{w_1' \cdot P_4^{IN} + w_2' \cdot P_4^{TD} + w_3' \cdot \text{Downsample}(P_3^{OUT})}{w_1' + w_2' + w_3' + \varepsilon}\right)\right) \quad (4)$$

where $Conv(\cdot)$ denotes a depth-wise separable convolutional operation, and $swish(\cdot)$ is the Swish activation function. Then, considering the semantic-spatial gap, we apply a resolution-aware fusion block (RFB) to fuse semantic and spatial information at the stage-level in the network, which takes the high-level features as the input and combines the low-level features. We generalize the fusion as follows, $F(x_l) = x_l \oplus x_l \otimes (x_{l-1} \oplus x_{l-1} \otimes Conv_{4\times4}(x_{l-2}))$, where $x_l$ is a local feature of the pyramid scale $l$, the sign $\otimes$ denotes element-wise multiplication, the sign $\oplus$ denotes element-wise addition, and $Conv_{4\times4}$ indicates a convolutional layer that containing $4\times4$ convolution kernels (here the stride set to 2, and padding set to 1). The RFB multiplies each element of the high-level and low-level features at their corresponding positions, and thus adds each element of deep feature and the feature after element-wise multiplication to the corresponding positions, which can preserve semantic-spatial features in bottom-up side pathway processes.

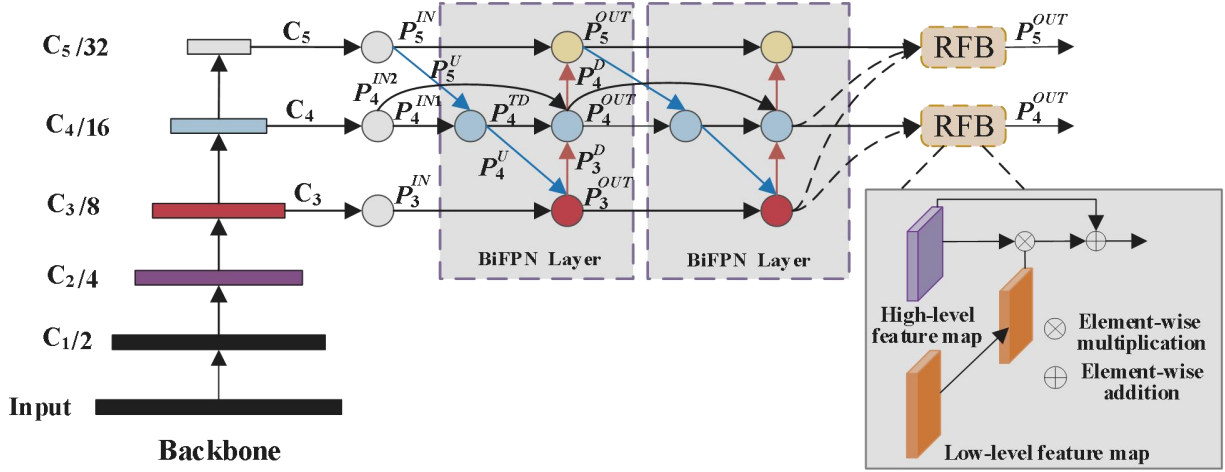

**Fig. 4** Illustration of pyramid-like semantic modulation block

### 3.3.2 Context-aware feature attention extraction

To better capture multi-scale and context-aware global information, we design a context-aware feature attention extraction (CFAE) module to encode input multi-scale features, as shown in Fig. 5. The CFAE module consists of the dense atrous spatial pyramid pooling (DenseASPP) and global context network (GCNet) blocks. The PSMB-extracted multi-scale convolutional response maps, $P_5^{OUT}$ and $P_4^{OUT}$, are fed into the CFAE module. Convolutional layers have local connectivity, shared weights, and long-range dependencies are extracted by stacking deep convolutional layers, however, these layers frequently cause gradient problems and optimization difficulties. These limitations can be alleviated using atrous convolutions and DenseASPP [41], which can avoid the loss of information in the pyramid pooling module (PPM) [42], and dense dilation convolutions, which enlarge the receptive field of kernels. We first apply the DenseASPP that contains a densely connected convolutional layer with different dilation rates 3, 5, and 7 to capture multi-receptive-field contextual information and 1×1 convolutional layer to incorporate the image-level features. The GCNet [43] module then employs a combination of non-local and SE blocks with $X \in \mathbb{R}^{C \times H \times W}$ as the GCNet input, where H×W is the size of the feature map, and C indicates the number of channels. The GCNet employs a squeeze-and-excitation module that establishes the inter-dependencies between channel-wise size, and spatial attention mechanism that establishes long-range dependencies, and it effectively models the global context information from a global perspective. In step (1), through a non-local block (i.e., simplified self-attention mechanism), the elements are multiplied separately to combine the features of all positions to form the global context feature $X \in \mathbb{R}^{C \times 1 \times 1}$, and the inter-dependencies between each channel are then captured through the SE block. In step (2), the transformation phase uses a point-wise convolution to reduce its dimensionality and obtain the feature maps $Y \in \mathbb{R}^{C/s \times 1 \times 1}$, where s is the reduction ratio. In step (3), the fusion module is added element-wise, and the global context features can be concatenated with the original feature maps. The output feature maps of the GCNet $\widetilde{X} \in \mathbb{R}^{C \times H \times W}$ are as follows:

$$\widetilde{X} = X + W_{v2}\left(LN\left(W_{v1}\left(\sum_{j=1}^{H \times W} \frac{exp(W_k x_j)}{\sum_{m=1}^{H \times W} exp(W_k x_m)} x_j\right)\right)\right) \qquad (5)$$

where LN indicates layer normalization, $x_j$ denotes the position of the query point in the feature maps, and $W_{v1}$, $W_{v2}$, and $W_k$ involve the feature weight vectors following a 1×1 point-wise convolution.

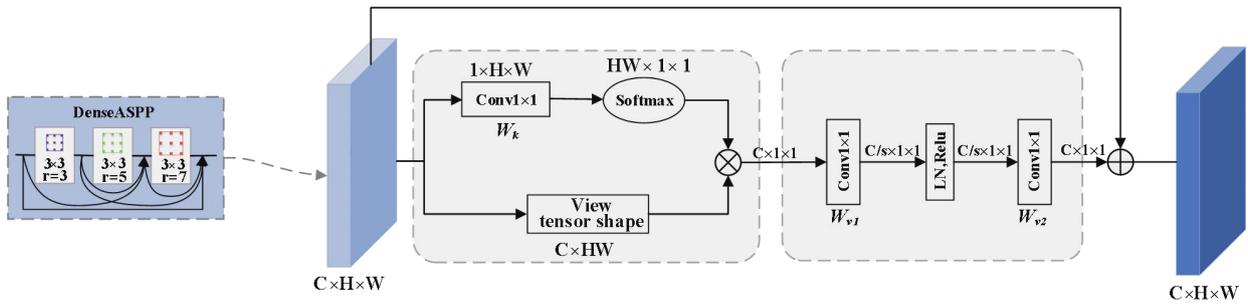

**Fig. 5** GCNet structure diagram

### 3.4 Joint attribute-based recognition

#### 3.4.1 Fashion attribute annotation

Clothing images contain rich visual information and multiple clothing attributes. We divide the clothing-rich annotations into two representative groups: PID labels and attribute labels. The clothing image PID labels can only define the distance between all clothing in an identity-related coarse-scale manner. The attribute labels explicitly guide the model to focus on local semantic information, which greatly converges the training of models. In addition, these discriminative visual representations can push clothing images with similar attribute labels closer to the feature space, and images with different attributes are farther in the feature space. The attributes will be utilized as a classification label to represent fashion collections, while the PID information can be utilized as a unique label for each fashion item separately. Concretely, we select 4 category attribute type labels from the total 18 category attribute value labels of the DeepFashion-C2S dataset. The detailed clothing attribute types are: sleeve-length (Slv-Len), type of collar (Collar), shell fabric of clothes (Fabric), fitness of clothes (Fitness). These attributes describe the characteristics of upper-wear, which can meet the needs that may arise when they perform a fine-grained search. In summary, four upper-wear attributes are created with 18 labels, as shown in Table 1. Fig. 6 illustrates the distribution of attribute items with the upper-wear genre on the DeepFashion C2S dataset.

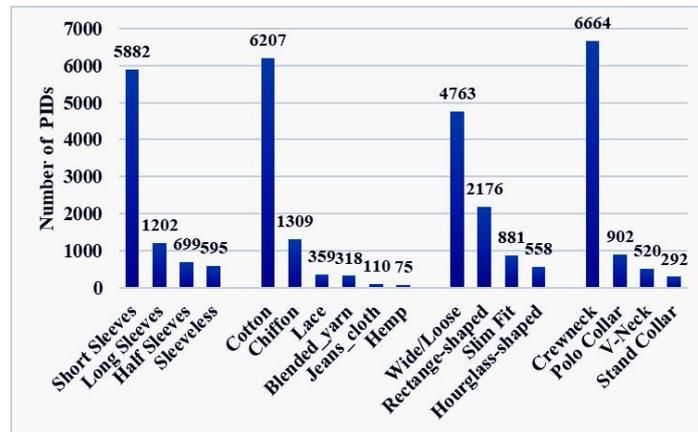

**Fig. 6** Distribution of attribute value with upper-wear genre. Each row shows the numbers of PIDs for attributes on DeepFashion C2S

**Table 1** The list of upper clothes attribute labels

| Attribute Type | Attribute Value |
| --- | --- |
| Slv-Len | Short Sleeves, Sleeveless, Half Sleeves, Long Sleeves |
| Collar | Crewneck, Polo Collar, Stand Collar, V-neck |
| Fabric | Cotton, Chiffon, Blended Yarn, Jeans Cloth, Lace, Hemp |
| Fitness | Wide/Loose, Slim Fit, Rectangle-shaped, Hourglass-shaped |

### 3.4.2 Multi-label classification with clothing attributes

Here, an image-level annotations supervised classification scheme is adopted for clothing attribute prediction. As shown in Fig. 7, a joint attribute-based recognition network (JARN) is introduced to learn a clothing image embedding and predict its attributes under a unified framework. JARN contains two prediction branches: clothing attribute prediction sub-task, and clothing identity classification sub-task. Let us assume we have a predefined list of attributes (e.g., Slv-Len, Collar, Fabric, and Fitness), which we index using the symbol $a$. $\mathcal{T} = \{(I_1, y_1, y_{a1}), \cdots, (I_N, y_N, y_{aN})\}$ is the fashion dataset with PIDs and attributes, where $I_i$ and $y_i$ denote the $i$-th clothing image and PIDs, $a_i$ denotes the $i$-th clothing attributes, and $N$ is the number of samples in the training set. Each attribute $y_a$ is associated with a list of possible attribute values, $y_{ai} = \{y_{ai}^1, y_{ai}^2, ..., y_{ai}^{J_a}\}$, where $y_{ai}^j$ is the $j$-th attribute label for the image $I_i$, and $J_a$ is the total number of possible values for that attribute. Inspired by Lu et al.[44], we utilize a fully feature-shared multi-task learning architecture for detailed attribute prediction by using an SFFP extractor to obtain CNN-extracted features. Then, image representation $f_n$ is fed into a fully connected two-layer network for each clothing attribute, which maps $f_n$ to attribute-specific embedding $v_{n,a} = \phi_a(f_n)$. Embedding $v_{n,a}$ is used to predict its attribute values for the given image via a classification layer, and the multi-layer fully connected layer yields logits which are passed through the log-softmax layer to obtain the prediction probability $\hat{y}_{n,a} = log\_softmax(v_{n,a})$. The attribute prediction layer loss function of clothing detail uses label smoothing regularization (LSR) loss for attribute prediction, where the number of classifiers equals the number of defined clothing attributes, and the number of fully connected neurons in each classifier equals the number of specific labels of clothing attributes corresponding to that classifier. For the PID classification, we assume that the embedding vector $f_n$ is fed into a fully connected two-layer fully-connected network via the classification of identity sub-tasks defined in the form of a cross-entropy loss.

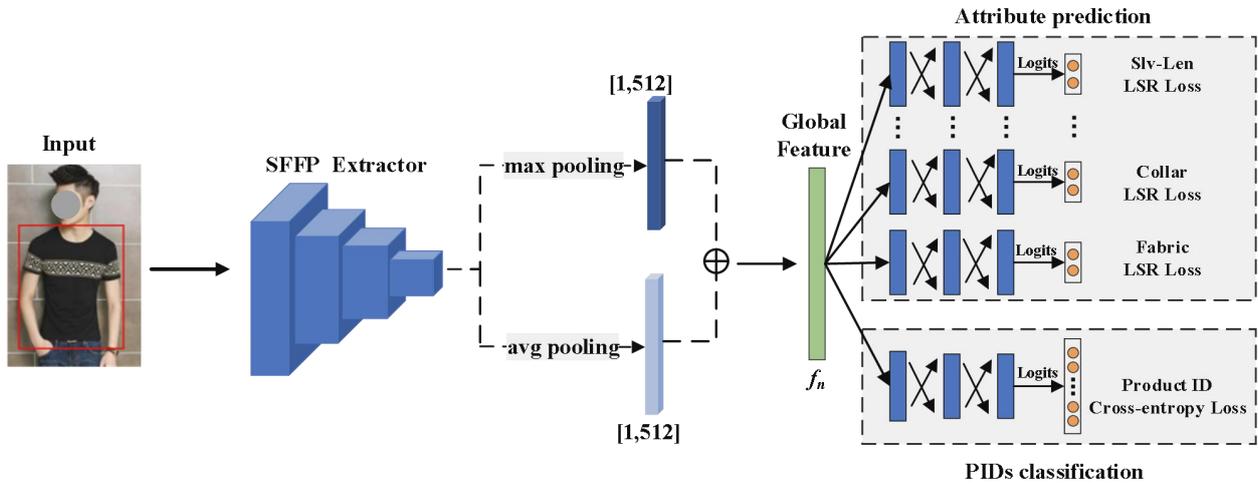

**Fig. 7** The JARN structure of our approach

### 3.5 Loss functions

#### 3.5.1 Metric loss learning with trihard loss and center loss

Triplet ranking loss and complex variants of triplet loss are widely utilized in image retrieval and ranking tasks to obtain a similarity metric. Clothing retrieval is considered to be a clustering or ranking task in feature leaning based on deep metric learning. Triplet loss is considered to optimize the intra-class compactness and the inter-class discriminability in the embedding space during the training. Cross-domain C2S clothing retrieval focuses on the similarities between different domains. In this study, a cross-domain triplet scheme is constructed to compensate for the domain gap.

Triplet ranking loss is trained using the distribution of triplets set $\mathcal{T} = \{(I_q, I_p, I_n)\}$, where $I_q$ denotes the anchor image, $I_p$ denotes a positive image, and $I_n$ denotes a negative sample. Following Sharma et al. [45], the sampling selection strategy $(I_q^C, I_p^S, I_n^S)$, $(I_q^C, I_p^C, I_n^C)$ and $(I_q^S, I_p^S, I_n^S)$ is adopted, where $\mathcal{D}^S$ and $\mathcal{D}^C$ denote images from the shop and consumer domains, respectively. A total of $(P+1) \times K$ images are sampled by constructing one mini-batch. K consumer images and K shop images are selected in the positive sample set, and $(P-1) \times K$ images are randomly selected in the negative sample set. The cross-domain sample triplet training set $\mathcal{T}$ is then constructed using mini-batch trihard loss for metric learning. Thus, mini-batch trihard sampling strategy can alleviate the problem of class-conflict. The specific implementation steps of mini-batch trihard loss as follow: In step (1), a cross-domain triplet set is used to perform hard-mining sample online, yielding the hardest positive and negative pairs. In step (2), the extracting features $F = [f^g, f^{P_h}, f_c^{P_h}, f^{P_v}, f_c^{P_v}, f^{L_a}]$ are concatenate and input into the trihard-based similarity metric learning and regularization term center loss. The trihard ranking loss function is formulated as follows,

$$\mathcal{L}_{\text{triplet}} = \sum_{i=1}^{P+1} \sum_{q=1}^{K} \left[ \overbrace{\max_{p=1...K}(D(F_q^{(i)}, F_p^{(i)}))}^{\text{hardest positive}} - \overbrace{\min_{\substack{n=1...K \\ j=1...P \\ j \neq i}}(D(F_q^{(i)}, F_n^{(j)}))}^{\text{hardest negative}} + m \right]_+ \quad (6)$$

where $[z]_+ = \max(z, 0)$, $F_p^{(i)}$ denotes the embedding feature of the *i-th* positive example, $F_n^{(j)}$ denotes the embedding feature of the *j-th* negative example, $D(\cdot, \cdot)$ denotes the Euclidean metric, and $m$ denotes a margin. To achieve training convergence, $L_2$-norm was performed on the embedding features before inputting them into the trihard loss function.

Following the previous work [11], the center loss was adopted as a penalty term with improved trihard loss as the auxiliary loss, which can reduce model overfitting to noisy samples. The center loss function is formulated as follows,

$$\mathcal{L}_{\text{center}} = \frac{1}{2} \sum_{j=1}^{M} \left\| F_j - c_{y_j} \right\|_2^2 \quad (7)$$

where $M$ denotes the batch size, $F_j$ denotes the *j-th* embedding feature, and $c_{y_j}$ denotes the center of the ground-truth label $y_j$ of the class feature.

### 3.5.2 Representation learning with multi-task classification loss

In addition to metric learning, label-smoothing regularization (LSR) [42] is implemented to optimize the calculation of multi-label attribute classification. This resolves the shortcomings of the conventional one-hot encoding (OHE) and prevents overfitting during feature training. In our LSR, $q'_k$ denotes the following smoothed label distribution,

$$q'_k = \begin{cases} 1 - \varepsilon + \dfrac{\varepsilon}{K} & \text{if } k = y_a \\ \varepsilon / K & \text{otherwise} \end{cases} \quad (8)$$

where $K$ is the total number of classes, $q'_k$ indicates the *k-th* ground truth attribute labels $\{y_a\}$, and $\varepsilon$ denotes the smoothing factor, with its value fixed at 0.1. The smoothing factor optimizes the LSR between the smoothed label distribution $q'_k$, and the outputs of the attribute prediction distribution $p_k$, which is defined as,

$$\mathcal{L}_{LSR} = -\sum_{k=1}^{K} q'_k \log(p_k) = (1 - \varepsilon) H(q, p) + \varepsilon H(u, p) \quad (9)$$

During PID classification training, the cross-entropy loss function is used to calculate the distance between the predicted labels and the ground-truth labels for PID classification,

$$\mathcal{L}_{CE} = -\frac{1}{N} \sum_{i=1}^{N} q_i \log(p_i) = \frac{1}{N} \sum_{i=1}^{N} \left( -z_i + \log\left( \sum_{j=1}^{K} \exp(z_j) \right) \right) \quad (10)$$

where $p_i$ is the probability of the i-*th* PID label of the softmax output, $q$ denotes the true distribution of each image, $z_j$ is the logits by feed forward propagation scheme. Hence, our MMFL-Net learns in a multi-task way via total loss $\mathcal{L}_{total}$, which can be expressed as follows,

$$\mathcal{L}_{total} = \mathcal{L}_{LSR} + \mathcal{L}_{CE} + \gamma_{triplet} \mathcal{L}_{triplet} + \beta_{center} \mathcal{L}_{center} \quad (11)$$

where $\gamma_{triplet}$ and $\beta_{center}$ are the trade-off hyperparameter to balance the $\mathcal{L}_{triplet}$ and $\mathcal{L}_{center}$ loss functions, respectively.

### 3.5.3 Training of proposed network

We have described the primary training procedure used to train and optimize our proposed network in Algorithm 1. The training involves of five steps. (1) Dataset acquisition and data loader construction. We initialize the setup train set data loader, $\mathbf{DL}_{train}$, using a triplet sampler; gallery set data loader, $\mathbf{DL}_{gallery}$; query set data loader, $\mathbf{DL}_{query}$ (lines 1–3). (2) Construct of a MMFL-Net network and initialization of our network with ImageNet pre-trained weights (lines 4–5). (3) Network feed forward propagation scheme. We compute the multi-task classification loss for multi-label attribute prediction and PID classification, metric loss, and overall loss of each mini-batch hard-mining sampling strategy using Eq. (11) (lines 9–11). (4) Network back-propagation scheme. We compute the back-propagation the gradients using Eq. (11), update the network weights with the Adam optimizer, update the center loss weights with the SGD optimizer, and update the learning rate of our proposed network using a multistep decay learning rate scheduler (lines 12–15). (5) Network evaluation test. We invoke the evaluation function that calculates mAP and Acc@K (lines 16–17).

---

**Algorithm 1:** The overall training procedure of MMSL-Net

---

**Input:** Training set $\mathcal{T} = \{(\mathbf{I}_i^T, \mathbf{y}_i^T, \mathbf{a}_i^T)\}$, gallery set $\mathcal{G} = \{(\mathbf{I}_i^G, \mathbf{y}_i^G)\}$, query set $\mathcal{Q} = \{\mathbf{I}_i^Q\}$, total training epochs $E$, batch size $B$, hyper-parameter $\gamma_{triplet}$ and $\beta_{center}$, the initial learning rate $\eta$, center loss learning rate $\mu$.

**Output:** Final MMFL-Net model parameters $\mathbf{\Theta}^N$, mAP and Acc@K.

1. $\mathbf{DL}_{train} \coloneqq$ set train set $\mathcal{T}$ data loader using triplet sampler;
2. $\mathbf{DL}_{gallery} \coloneqq$ set gallery set $\mathcal{G}$ data loader;
3. $\mathbf{DL}_{query} \coloneqq$ set query set $\mathcal{Q}$ data loader;
4. $\mathcal{F} \coloneqq$ set our MMFL-Net network;
5. $\mathbf{\Theta}^0 \leftarrow$ Initialize our model weight with the ImageNet-pretrained IBN-Net;
6. **for** $e \leftarrow 1$ **to** $E$ **do**
7.     **for** sampled mini-batch $\mathcal{B} \in \mathbf{DL}_{train}$ **do**
8.         Obtain feedforward scheme predicted logits $v_{n,a}, v_{n,id}$ and extracting feature $F$;
9.         Calculate the multi-task classification loss of $v_{n,a}, v_{n,id}$ by Eq. (9) and Eq. (10) with the mixup;
10.         Calculate the deep metric loss of $F$ by Eq. (6) and Eq. (7);
11.         Calculate the total training loss by Eq. (11);
12.         Calculate the back-propagation the gradients $\nabla \mathcal{L}(\theta^e)$ using Eq. (11);
13.         Update the network weight parameters by Adam optimizer,

$$\theta^{e+1} \leftarrow \theta^e - \frac{\eta}{\sqrt{\beta_2 S^e + (1-\beta_2)\nabla\mathcal{L}(\theta^e)\cdot\nabla\mathcal{L}(\theta^e)} + \alpha}(\beta_1 v^e + (1-\beta_1)\nabla\mathcal{L}(\theta^e))$$

13.         Update center loss weight parameters by SGD optimizer, $\theta_c^{e+1} \leftarrow \theta_c^e - u\nabla\mathcal{L}_{center}(\theta_c^e)$
14.     **end for**
15.     Update the learning rate $\eta$ using a multistep decay learning rate scheduler;
16.     Invoke the evaluation function,
17.         mAP, Acc@K $\leftarrow$ R1_mAP_eval($\mathbf{DL}_{gallery}, \mathbf{DL}_{query}, \mathcal{F}, \mathbf{\Theta}^N$)
18. **end for**
19. **Return:** $\mathbf{\Theta}^N$, mAP, Acc@K

# 4. Experimental results

## 4.1 Experimental setup

### 4.1.1 Datasets and evaluation protocol

**DeepFashion C2S (DF-C2S) dataset**. This is a large-scale clothing image dataset and was originally introduced by Liu et al. [8], which consists of over 800,000 images and several fashion-related benchmarks. Each image has very rich annotations, including 50 categories, fine-grained attribute annotations, bounding boxes (bbox) annotations, parsing mask annotations, fashion landmark annotations, and about 300,000 pairs of images of different poses and scenes. The data that were used in this study as the C2S clothing retrieval benchmark that consists of 33,881 unique clothing items with 239,557 images, and 195,540 C2S pairs. Each image is accompanied by 303 attribute value annotations, bbox annotations, and fashion landmarks. These include categories such as clothing, dresses, tops and trousers; the training dataset also includes both coarse and fine granularity.

**Street2Shop (S2S) dataset.** This dataset is the first large-scale fashion datasets introduced by Kiapour et al. [6]. It has become one of the most widely used datasets for evaluating cross-domain fashion retrieval tasks. Street2Shop has eleven different categories of fashion images: bags, belts, dresses, eyewear, footwear, hats, leggings, outerwear, pants, skirts, and tops. The benchmark contains 204,795 distinct fashion items with 404,683 online shop images, and 20,357 street snap-shot images. The online shop images do not provide bbox annotations, so the un-cropped shop images are used as gallery images. Five categories, including tops, dresses, skirts, pants, and outerwear were selected for this study. The images of each category contained two domains; the street domain was divided into two parts, the training and test sets.

**Evaluation protocol.** To evaluate how well our proposed model performed in cross-domain C2S fashion retrieval, we report the cumulative matching characteristics (CMC) at Top-K retrieval accuracy (Acc@K) and mean average precision (mAP) for evaluation on both datasets. All the experiments are performed in single query and multi-shot episodes.

The size of the above two datasets is significant. To improve the retrieval accuracy of our model and reduce the computational cost, we construct our DF-C2S and Street2Shop sub-datasets, with which the number of images and class specificity are reduced. Our C2S sub-datasets have the following characteristics. (1) The data subset includes the refine shop and unrestricted consumer images and uses single-subject (single-model or tiled single-piece) clothing images. (2) The images are annotated with rich annotation information regarding the clothing items, including attribute types, attribute values, bbox annotations, and fashion landmarks. (3) The attribute values under a specific attribute type are mutually exclusive, and the attribute values under different attribute types are independent of each other. For example, a V-neck and crewneck cannot co-exist in the same image in the collar-type attribute. Notably, we dropped some specific images during the dataset selection to guarantee the mutual exclusivity of attributes. Table 2 summarizes the data distribution characteristics of DF-C2S and S2S.

Table 2. The data characteristics of DF-C2S and S2S

| Dataset | Images | Training | Image data | Product ID | ClothItems | AttrValues | Location |
|---|---|---|---|---|---|---|---|
| DF-C2S | 57,812 | 43,654 | Model/Tiled | 6,063/2,073 | 5 | 18 | bbox |
| S2S | 47,646 | 36,947 | Model/Tiled | 3,005/1,688 | 5 | N/A | bbox w/o shop |

### 4.1.2 Implementation details

The proposed multi-branch and multi-task network and both losses are implemented using the PyTorch framework with NVIDIA Tesla V100 GPU of 32GB memory. The proposed model adopts the ResNet50-IBN-A backbone initialized with ImageNet pre-trained weights. We utilize YACS, a configuration management system, to define all the hyper-parameters used in the CBIR pipeline. To avoid the image distortion, the longest edges of the images are first scaled to 320, and the short edges are padded with pixels along both sides of the images in line with the longest edges, while keeping the aspect ratio and resizing to 320×320 pixels before being fed into the network. During training, we augment the samples by random horizontal flipping, rotating, random center cropping, color jittering, and multi-data mixup [46], weighted by averaging of different samples. Our model is trained by mini-batch with randomly triplet sampled images. The batch size is 32=(2×(P+1)×K), which contains P identities, and each identity has 4 images, i.e., P set to 3 and K set to 4. The max number of epochs is set to 120. We use the Adam optimizer for fast and robust back propagation and loss convergence, with the weight decay factor is set to 5e-4. The initial learning rate is set as 0.0001 with a multistep learning rate scheduler (default hyperparameter settings: $\beta_1 = 0.9$, $\beta_2 = 0.999$, $\alpha = $ 1e-8), which decrease the learning rate decay at 50-th and 100-th training epochs by a factor of 10. An SGD optimizer with momentum 0.9 separately optimizes the center loss with a learning rate set to 0.5. All of the $1\times1$ convolutions use Kaiming initialization with a normal distribution. In Eq. (8), $m$ is empirically set to 0.3. In Eq. (11), we empirically set the total loss functions balance weights $\beta_{center}$ to 0.0005 and $\gamma_{triplet}$ to 1.5.

### 4.1.3 Parameter influence

In the training phase, the model is trained and optimized by total loss, as defined by Eq. (11). Thus, the margin parameter $m$ and the weight parameter $\gamma_{triplet}$ can affect the similarity metric learning performance for trihard ranking loss in the feature embedding space. In particular, $m$ denotes the threshold between the relative distance of the query to the positive image and the relative distance of query to the negative image, and $\gamma_{triplet}$ is the trade-off hyperparameter to balance the $\mathcal{L}_{triplet}$ loss functions. To study the impact of these parameters on the overall performance of the model, we conduct experiments on the DF-C2S datasets and evaluate the retrieval performance with mAP and Acc@K.

To study the impact of weight parameter $\gamma_{triplet}$, we set the margin parameter $m$ as 0.3 and then assign incremental values from 2.0 to 1.0 in steps of 0.5 for $\gamma_{triplet}$. The experimental results show that have a slight discrepancy in mAP value, but when $\gamma_{triplet}$ is 1.5, Acc@50 is 85.0%, and equivalent to that of the state-of-the-art method. Additionally, it exceeds the current best method in all evaluation protocols (mAP, Acc@K) of retrieval performance.

To study the impact of the margin parameter $m$, we set the weight parameter $\gamma_{triplet}$ as 1.5 and then set $m$ from 0.2 to 0.6. The experimental results are presented in Table 3. For the margin parameter $m$, we observe that the evaluation retrieval performance fluctuates as the value of $m$ increases. By analyzing of mAP and Acc@1,10, we observe that the optimal value for $m$ is 0.3. Finally, we set $m$ as 0.3 and $\gamma_{triplet}$ as 1.5 for the following experimental evaluations on the two datasets, DF-C2S and S2S.

Table 3. Influence of the margin parameter $m$ via a fixed-parameter $\gamma_{triplet}$ on DF-C2S

| $\gamma_{triplet}$ | $m$ | mAP | Acc@1 | Acc@10 |
| --- | --- | --- | --- | --- |
| 1.5 | 0.2 | 52.1 | 48.0 | 71.0 |
| 1.5 | 0.3 | 53.4 | 48.7 | 72.2 |
| 1.5 | 0.4 | 52.2 | 47.9 | 71.4 |
| 1.5 | 0.5 | 53.0 | 48.8 | 71.9 |
| 1.5 | 0.6 | 52.1 | 47.2 | 71.6 |

## 4.2 Experimental results

### 4.2.1 Quantitative results

**Results on DeepFashion-C2S.** To evaluate and compare the performance of our proposed network, the quantitative results obtained using the state-of-the-art methods on the DeepFashion-C2S dataset are listed in Table 4. We evaluate them in terms of mAP and Top-K retrieval accuracy (Acc@1, 20, 50). The best performance achieves 53.4% mAP and 48.7% Acc@1 accuracy, which is superior over the state-of-the-art methods. Compared with CTL-L-R50IBN-CE (the current state-of-the-art method), our model achieves a +4.2% higher mAP, +11.4% higher Acc@1, and +1.0% higher

Acc@20. Our model outperforms L-R50IBN (a suboptimal state-of-the-art method) by a +10.9% higher Acc@1 and +10.4% higher mAP. Additionally, we use the k-reciprocal re-ranking post-processing method [47] by comparing with L-R50IBN(re-ranking)[11], which improves the accuracy of mAP and Acc@1 (+10.6%, +12.6%), respectively.

**Table 4.** Comparison of retrieval accuracy with the state-of-the-art methods on DeepFashion-C2S

| Methods | Backbone | Image Size | mAP | Acc@1 | Acc@20 | Acc@50 |
|---|---|---|---|---|---|---|
| FashionNet [8] | VGG-16 | 224 × 224 | - | 7.4 | 18.8 | 22.6 |
| VAM+ImgDrop [30] | GoogleNet | 224 × 224 | - | 13.7 | 43.9 | 56.9 |
| GSN [48] | Inception-V1 | 227 × 227 | - | 25.0 | 47.0 | 57.0 |
| GRNet [9] | GoogleNet | 224 × 224 | - | 25.7 | 64.4 | 75.0 |
| UDMA-E2E [45] | ResNet-101 | 224 × 224 | 25.5 | 37.8 | 55.1 | 65.2 |
| S-R50 [11] | ResNet-50 | 256 × 128 | 32.4 | 28.1 | 65.5 | 74.2 |
| M-R50IBN [11] | ResNet50-IBN-A | 224 × 224 | 40.5 | 35.8 | 75.1 | 82.4 |
| L-R50IBN [11] | ResNet50-IBN-A | 320 × 320 | 43.0 | 37.8 | 77.2 | 84.1 |
| CTL-L-R50IBN-CE [49] | ResNet50-IBN-A | 320 × 320 | 49.2 | 37.3 | 77.7 | **85.0** |
| MMFL-Net | ResNet50-IBN-A | 224 × 224 | 50.0 | 46.1 | 76.1 | 82.7 |
| **MMFL-Net (ours)** | ResNet50-IBN-A | 320 × 320 | **53.4** | **48.7** | **78.7** | **85.0** |
| L-R50IBN(re-ranking) [11] | ResNet50-IBN-A | 320 × 320 | 47.3 | 40.0 | 79.0 | - |
| **MMFL-Net (re-ranking)** | ResNet50-IBN-A | 320 × 320 | **57.9** | **52.6** | **79.5** | **85.5** |

**Notes**: *The best performance for each metric is presented in bold. S, M, and L in the model name indicates input image size. R50 or R50IBN indicates which backbone CNN was used, ResNet-50 or ResNet50-IBN-A, respectively. CE denotes centroid-based evaluation.*

**Results on Street2Shop.** We further compare our model with the state-of-the-art results on the Street2Shop dataset, and the results are shown in Tables 5 and 6. Table 5 shows the five categories retrieval performance of our approach and [11, 19] on the Street2Shop dataset based on five categories. In the dresses and skirts category, our model needs to improve in re-ranking the retrieval results. Its poor performance may be due to broken download links of some images in the dataset. Additionally, the unconstrained evaluation reported by Wieczorek et al. [11] as overall performance. As shown in Table 6, our model without re-ranking achieves Acc@1, Acc@10, and Acc@20 metrics of 57.7%, 77.8%, and 83.2%, respectively. Compared with the Wieczorek et al. [49], our model achieves +4.0% higher Acc@1, +6.9% higher Acc@10, +8.2% higher Acc@20. The performance is reported in terms of mAP, which is also +3.8% higher than Wieczorek et al. [11]. Additionally, it is found that our model outperforms re-ranking by +1.6% and +2.7% in terms of Acc@1 and mAP, respectively.

**Table 5.** Comparison of retrieval accuracy with the state-of-the-art methods on five categories of Street2Shop

| Category | Metrics | Kucer et al.[19] | L-R50IBN[11] | L-R50IBN†[11] | MMFL-Net | MMFL-Net† |
|---|---|---|---|---|---|---|
| Tops | mAP | 36.9 | 46.8 | 56.5 | **51.9** | **57.9** |
| | Acc@1 | 42.7 | 52.7 | **57.9** | 54.4 | 57.7 |
| | Acc@20 | 61.6 | 71.9 | 72.9 | **83.2** | 79.8 |
| Dresses | mAP | 49.5 | 65.8 | **75.3** | 67.9 | 73.7 |
| | Acc@1 | 56.4 | **73.7** | 76.7 | 73.6 | 73.9 |
| | Acc@20 | 72.0 | 85.9 | 85.8 | **90.1** | 88.5 |
| Skirts | mAP | 62.3 | 64.5 | 73.3 | 67.5 | **73.3** |
| | Acc@1 | 68.0 | 71.2 | **75.1** | 71.8 | 73.2 |
| | Acc@20 | 80.2 | 86.5 | 85.5 | **91.8** | 92.1 |
| Pants | mAP | 28.2 | 23.9 | 30.4 | **44.9** | 45.1 |
| | Acc@1 | 33.3 | 27.3 | 33.3 | **44.1** | 40.7 |
| | Acc@20 | 51.5 | 42.4 | 48.5 | **64.4** | 67.7 |
| Outwear | mAP | 28.1 | 36.8 | 45.3 | **47.1** | **51.9** |
| | Acc@1 | 30.5 | 43.5 | **51.9** | **50.0** | 51.5 |
| | Acc@20 | 47.9 | 68.0 | 70.2 | **81.3** | **81.7** |

Notes: *The best performance for each metric is presented in bold.* † *denotes model with re-ranking.*

Table 6. Comparison of unconstrained retrieval results with the state-of-the-art methods for the Street2Shop

| Method | mAP | Acc@1 | Acc@10 | Acc@20 | Acc@50 |
|---|---|---|---|---|---|
| Kucer et al. [19] | 29.7 | 34.4 | - | 60.4 | - |
| M-R50IBN [11] | 42.4 | 49.2 | 66.2 | 70.9 | - |
| L-R50IBN [11] | 46.8 | 53.7 | 69.8 | 73.6 | - |
| CTL-L-R50IBN [49] | 45.9 | 53.3 | 68.9 | 72.8 | 78.2 |
| CTL-S-R50-CE [49] | 49.8 | 43.2 | 61.9 | 66.0 | 72.1 |
| CTL-L-R50IBN-CE [49] | **59.8** | 53.7 | 70.9 | 75.0 | 79.2 |
| **MMFL-Net (ours)** | 50.6 | **57.7** | **77.8** | **83.2** | **89.3** |
| L-R50IBN (re-ranking)[11] | 54.8 | 57.1 | 72.9 | - | - |
| **MMFL-Net (re-ranking)** | **57.5** | **58.7** | **75.7** | **81.0** | **88.1** |

### 4.2.2 Qualitative results

To show the reasoning effect of our proposed model more intuitively, we select some representation query images and visualize the retrieval results for clothing images. The experimental results of our network on DeepFashion-C2S and Street2Shop datasets are shown in Fig. 8 and Fig. 9, respectively. We implement a web visual search platform based on Flask and Vue CLI for practical applications. The retrieval results with a red tag are an incorrect match to the query image, while the ones with a green tag are the correct retrieval results. We use the continuous feature vector extractor network to extract image features from the offline dataset and build an all-features matrix database; then, we create an index image path and the corresponding feature vectors in the matrix. We apply K-means clustering algorithm to cluster the feature vectors of images from the shop domain in the retrieved gallery set to speed up the retrieval process. Interestingly, the visualization results in the last row of Fig. 8 shows that the same clothing items is retrieved for different PIDs with visually similar contents. It shows that MMFL-Net learns the multi-scale and multi-granularity of features through training. We select images with challenging user queries, such as from different viewing angles, with image rotation, poor lighting, and image scale changes. We obtain better retrieval performance using our MMFL-Net on the cross-domain clothing images datasets.

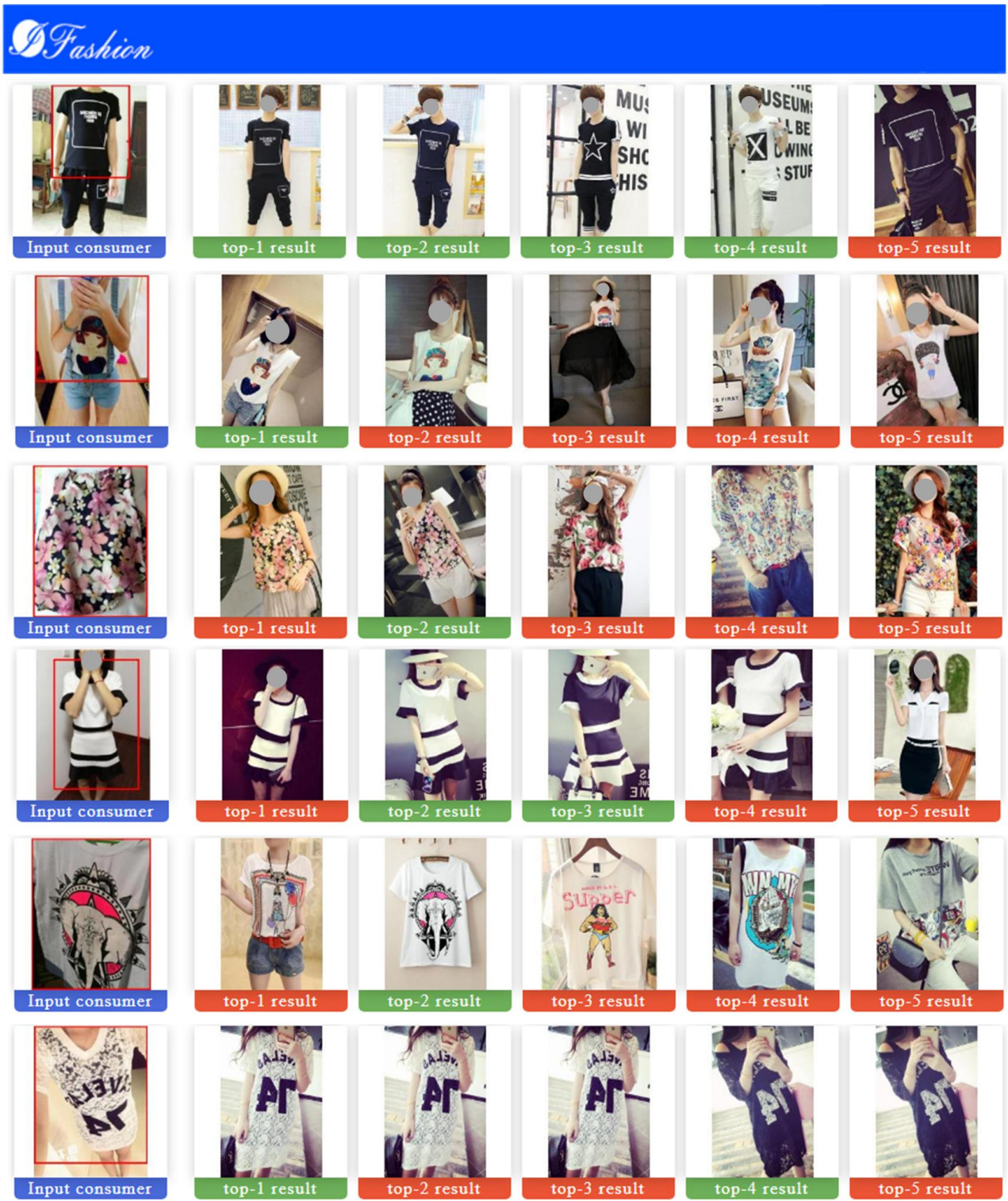

**Fig. 8** Top-5 ranked retrieval results on DeepFashion-C2S dataset. Query images with blue tag, the retrieval images with green tag are the correct match to the query, while red tag are the incorrect matching.

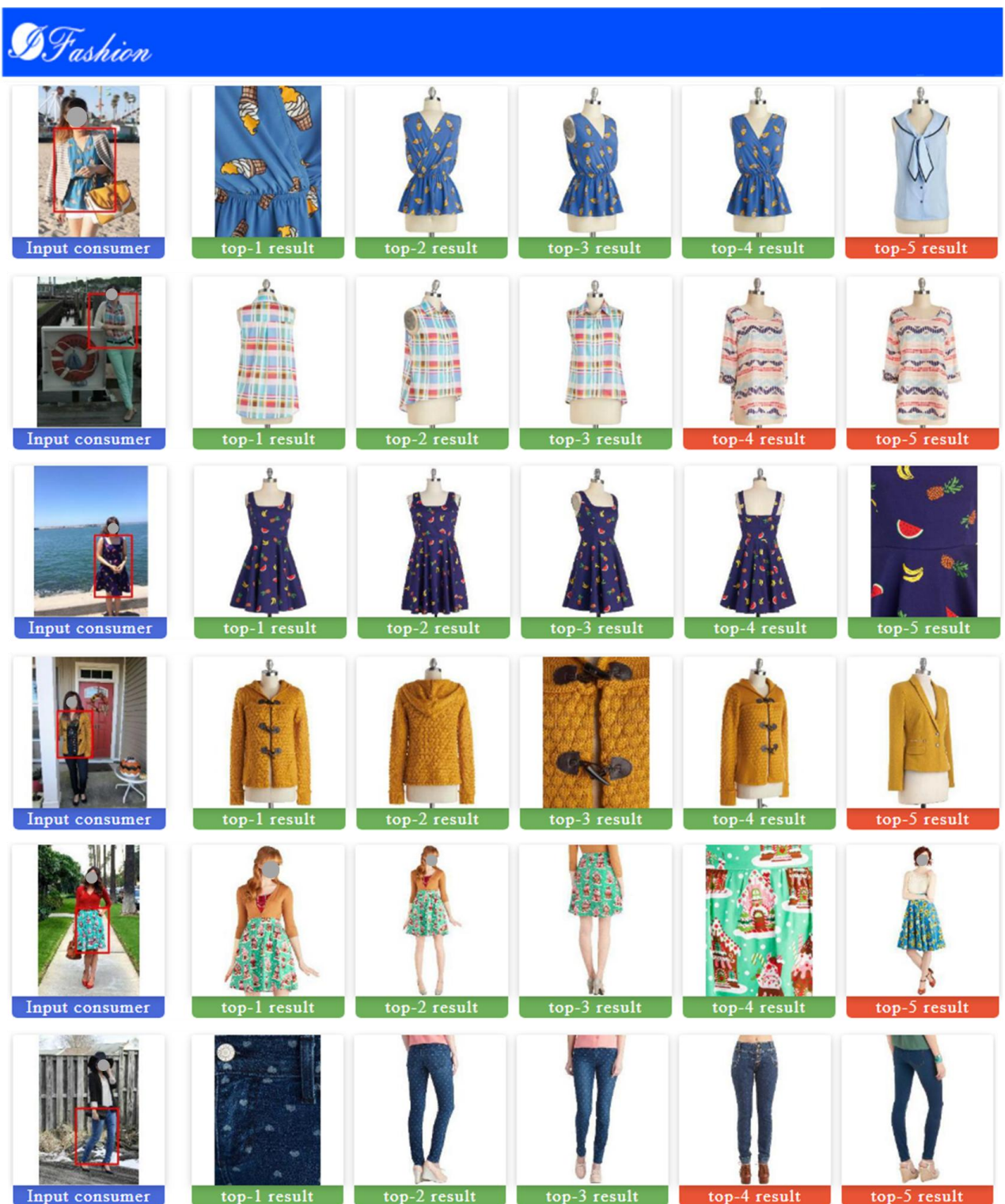

**Fig. 9** Top-5 retrieval results on Street2shop dataset. Blue tag denotes query; Green tag denotes correct; Red tag denotes incorrect.

#### 4.2.3 Results of attribute prediction

For the model training and evaluation of clothing attribute prediction, 57,812 clothing images are selected from the DeepFashion-C2S dataset according to the pre-defined upper-wear attribute labels in Table 1. We select 43,654 images from the evaluation status (train/validation) of the evaluation partition file as the training set, and 14,158 images from the evaluation status (test) of the evaluation partition file as the test set. Fig. 10 shows the confusion matrix for multi-label classification of the four attributes in the test set, where each column indicates the prediction results of attribute type, and each row indicates the actual attribute type. For attribute prediction, we employ the Top-K prediction accuracy as the evaluation metric. The classification accuracy (Top-1 accuracy) of Slv_Len label is 79.93%, that of Collar label is 86.08%, that of Fitness label is 54.35%, and that of Fabric label is 79.33%, and our Top-1 accuracy is significantly better than those of MTCT[50] and CNN-MT[51]. We exploit the proposed JARN structure; the overall precision of the classification of upper-wear clothing attributes Slv_Len, Collar, Fitness and Fabric type is 63.55%, 62.72%, 40.75%, 45.65%, and the overall recall of 57.08%, 55.5%, 38.17%, 38.9%, and the F1-score is 60.14%, 58.89%, 39.42%, 42.01%, respectively.

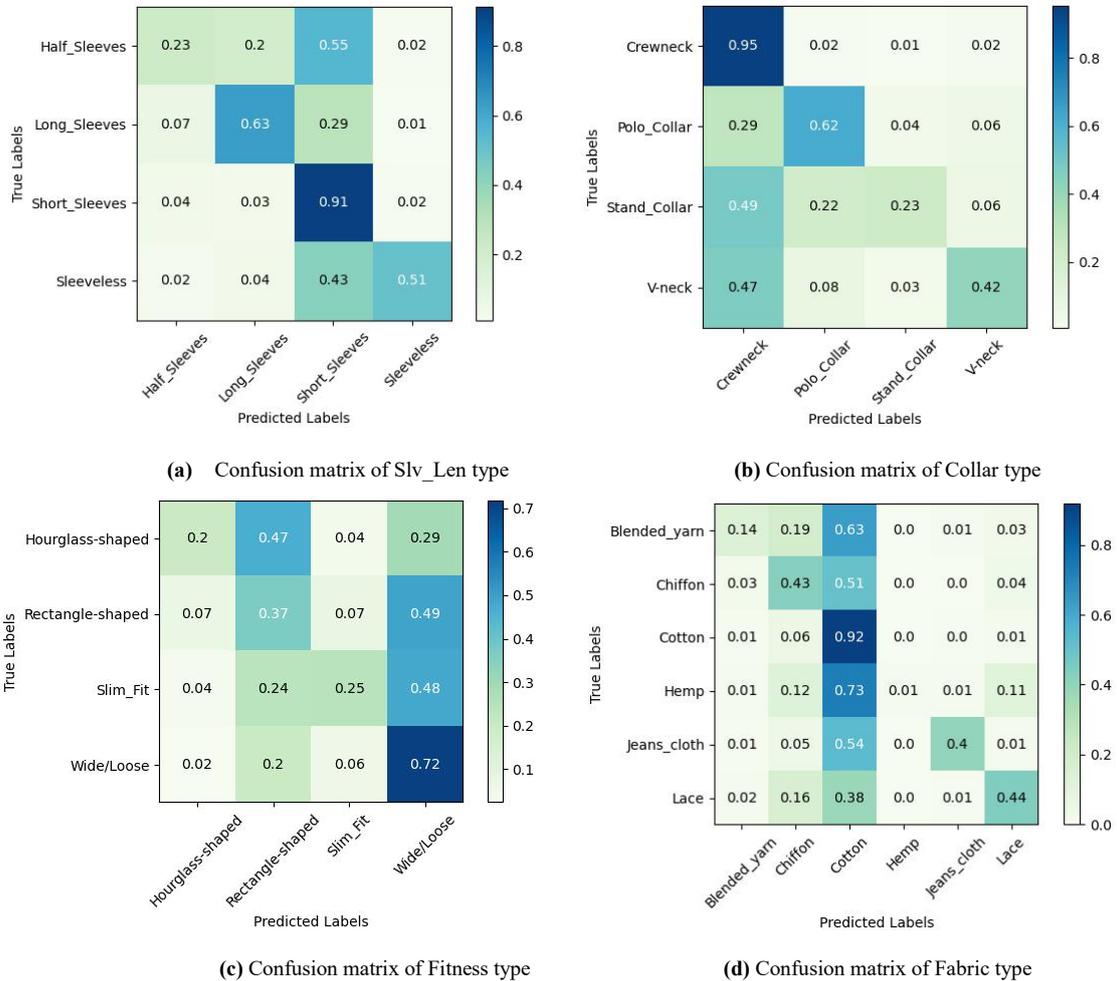

(a) Confusion matrix of Slv_Len type  (b) Confusion matrix of Collar type

(c) Confusion matrix of Fitness type  (d) Confusion matrix of Fabric type

**Fig.10** Confusion matrix of attribute multi-label classification. (a) Confusion matrix of Slv_Len type, (b)Confusion matrix of Collar type, (c) Confusion matrix of Fitness type, (d) Confusion matrix of Fabric type

### 4.3 Ablation studies

#### 4.3.1 Ablation experiment of different modules

To evaluate the viability of each module in the MMFL-Net, we conduct extensive ablation studies by deliberately removing specific components and comparing the evaluation results in a single query mode. Table 7 lists the performance of each module on DeepFashion-C2S and Street2Shop datasets. The ablation experiment is conducted on a baseline based on the ResNet-50 backbone, along with using LSR loss, cross-entropy loss, trihard loss, and center loss, which achieves 46.9% mAP and 41.4% Acc@1 accuracy on the DeepFashion-C2S dataset, and 45.4% mAP and 46.2% Acc@1 accuracy on the Street2Shop dataset. As each component is incrementally added, the retrieval accuracy improves, which justifies the performance of our proposed multi-branch network architecture. Especially, the GF-branch and PHF-branch boost Acc@1 and mAP by +6.2% and +5.2%, respectively, on DeepFashion-C2S, and by +8.0% and +2.8%, respectively, on Street2Shop. On this basis, we add the PVF-branch component, and the performance gains on the two datasets are (+0.4%, +0.4%) and (+2.3%, +1.5%) in terms of Acc@1 and mAP, respectively. Finally, we add our LDF-branch component, and obtain an improvement of (+0.7%, +0.9%) and (+1.2%,

+0.9%) in terms of Acc@1 and mAP on the two datasets, respectively. It can be seen that the PHF-branch/PVF-branch and LDF-branch are effective in improving the accuracy of our model. Our proposed network can push embedding learning to combine coarse-scale global information and part-informed local detailed information for a more powerful visual representation, thereby enhancing the discrimination of the different category instances.

**Table 7.** Performance with each module of our proposed network on DeepFashion-C2S and Street2Shop

| Method | DeepFashion-C2S | | | | Street2Shop | | | |
|---|---|---|---|---|---|---|---|---|
| | mAP | Acc@1 | Acc@10 | Acc@20 | mAP | Acc@1 | Acc@10 | Acc@20 |
| Baseline (ResNet-50) | 46.9 | 41.4 | 67.9 | 73.2 | 45.4 | 46.2 | 66.6 | 74.2 |
| GF-branch | 48.3 | 43.6 | 67.8 | 74.8 | 40.5 | 45.9 | 70.2 | 76.9 |
| + PHF-branch | 52.1 | 47.6 | 70.5 | 77.6 | 48.2 | 54.2 | 77.0 | 82.4 |
| + PVF-branch | 52.5 | 48.0 | 71.3 | 78.2 | 49.7 | 56.5 | 77.7 | 82.9 |
| + LAF-branch | 53.4 | 48.7 | 72.2 | 78.7 | 50.6 | 57.7 | 77.8 | 83.2 |

### 4.3.2 Comparison of different backbones

To evaluate the justifiability of our proposed framework, we conduct several experiments on different backbones using the DeepFashion-C2S benchmark, and the results are shown in Table 8. The first rows of Table 8 show the results of the SFFP structure. The backbone of our network uses different variants of the ResNet50 architectures: ResNet-50, SeResNet-50, Res2Net50, and ResNet50-IBN-A. ResNet50-IBN-A performs the best in terms of accuracy, and the ResNet50-IBN-A backbone with instance normalization may be learn more domain-invariant features. Compared with the Res2Net50 backbone network, our model achieves +1.5% higher mAP and +1.3% higher Acc@1. Additionally, the total number of parameters of MMFL-Net is 37.33M, and its FLOPs is approximately 64.23G. Compared with other backbone networks, the backbone of our model has fewer parameters and FLOPs.

**Table 8.** Comparison of different backbone on DeepFashion-C2S

| Backbone | Parameters | FLOPs | DeepFashion-C2S | |
|---|---|---|---|---|
| | | | mAP | Acc@1 |
| **ResNet50-IBN-A + PSMB + CAFE (ours)** | 37.33 M | 64.23 G | **53.4** | **48.7** |
| ResNet-50 + PSMB + CAFE | 37.33 M | 64.23 G | 49.0 | 44.9 |
| SeResNet50 + PSMB + CAFE | 39.87 M | 62.13 G | 51.1 | 47.1 |
| Res2Net50 + PSMB + CAFE | 37.48 M | 65.78 G | 51.9 | 47.4 |

### 4.3.3 Comparison of different losses

To evaluate the effectiveness of the improved hard mining trihard ranking loss, we conduct experiments with several relevant metrics and classification loss functions designed for the proposed MMFL-Net architecture. Previous work [8] has modeled joint training for a clothing retrieval task with classification loss and metric loss. In this paper, we compare with only the loss functions are replaced. The trihard, triplet-center loss functions are listed in Table 9. The trihard loss and cross-entropy loss results exhibited mAP and Acc@1 accuracy of 53.4% and 48.7%, respectively. The performance of the re-implemented trihard loss are observed to be consistently better than those of the triplet-center loss. In addition, we compare it to the performance of classification loss, when cross-entropy loss was replaced with L-Softmax loss, mAP and Acc@1 decrease by 4.6% and 4.4%, respectively; and when it is replaced with ArcFace loss, they decrease 18.4% and 18.2%, respectively.

**Table 9.** Comparison of different loss functions on DeepFashion-C2S

| Loss functions | | | | | DeepFashion-C2S | | | |
|---|---|---|---|---|---|---|---|---|
| trihard | triplet-center | cross-entropy | L-Softmax | ArcFace | mAP | Acc@1 | Acc@10 | Acc@20 |
| ✓ | ✗ | ✓ | ✗ | ✗ | **53.4** | **48.7** | **72.2** | **78.7** |
| ✗ | ✓ | ✓ | ✗ | ✗ | 50.8 | 45.9 | 68.9 | 74.0 |
| ✓ | ✗ | ✗ | ✓ | ✗ | 48.8 | 44.3 | 67.4 | 73.6 |
| ✓ | ✗ | ✗ | ✗ | ✓ | 35.0 | 30.5 | 52.8 | 60.3 |

**Notes:** *The symbols ✓ and ✗ denote that the corresponding information is included or excluded.*

### 4.4 Grad-CAM visualization attention maps

To investigate our proposed MMFL-Net, we employ the gradient-weighted class activation maps (Grad-cam) [52] to visualize the ROI regions to evaluate whether the granularity of features is discriminative. As shown in Fig. 11, the learned Grad-cam obtain a relatively high response from GF-branch, PHF-branch, PVF-branch, and LAF-branch. The GF-branch focuses on global salient areas, which can also increase the risk of network overfitting; the PHF-branch focuses on the part-level local area near the collar; the PVF-branch focuses on the part-level local region near the sleeve; and the LDF-branch focuses on multiple clothing components and learns more fine-grained discriminative features. These features can better distinguish different clothing images and lead to significant performance improvements.

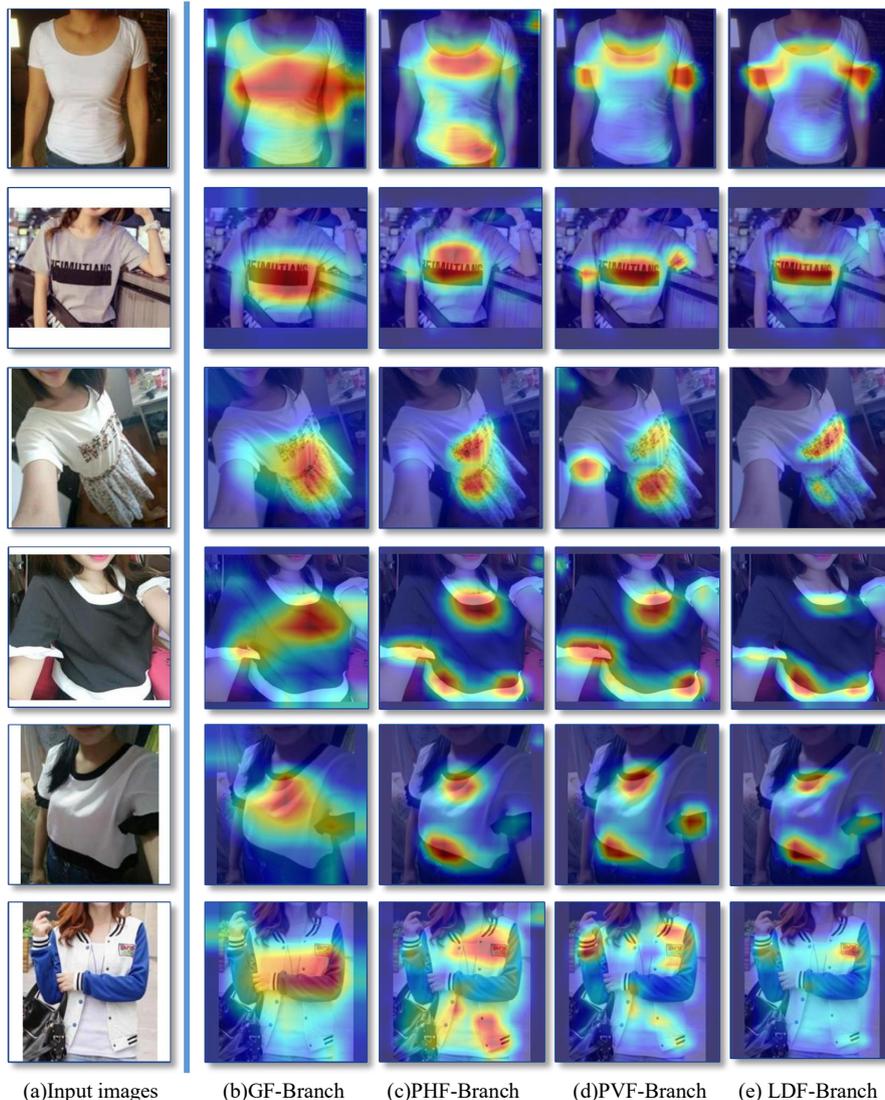

(a)Input images   (b)GF-Branch   (c)PHF-Branch   (d)PVF-Branch   (e) LDF-Branch

**Fig.11** Gradient-weighed class activation mapping (Grad-CAM) visualization of input sample. (a) Input images, (b) Results of the GF-Branch, (c)Results of the PHF-Branch, (d) Results of the PVF-Branch, (e) Results of the LDF-Branch.

## 4.5 Analysis of complexity and efficiency test

### 4.5.1 Complexity analysis

For the model complexity analysis of our proposed network, we obtained the model size, parameters, computing complexity FLOPs, inference time, and throughput at the inference stage. Specially, we measure the number of parameters and FLOPs our proposed network requires to feed forward propagation on a given image via an open-source toolbox (ptflops). The memory of the model size is 286.45MB, the total number of parameters is 37.33M, and its FLOPs is approximately 64.23G. Additionally, we also measure the inference time and throughput of our model. Note that the inference time and throughput are recorded on an NVIDIA Tesla 32G-V100 GPU with RGB input of $3 \times 320 \times 320$. The inference time is approximately 27.1ms, and average FPS is 36.8/s with a batch size of 1. The throughput time is approximately 168.2 img/s with a batch size of 64.

### 4.5.2 Efficiency test

We evaluate the training and retrieval inference time of our proposed model. The experiment is conducted on a normal computer with 47G RAM and NVIDIA Tesla 32G-V100 GPU with RGB input of $3 \times 320 \times 320$. For training time, our model takes about 68h and 50h on DF-C2S and S2S, respectively. During the indexing phase, the accuracy of our model decreases significantly when PCA dimensionality reduction and whitening are applied simultaneously. Furthermore, the accuracy does not improve by applying the PCA dimensionality reduction either. It is, therefore, determined that our model would directly index the 3072-dim feature embeddings extracted through the network. To obtain the retrieval time efficiency, our model uses the k-means clustering method to cluster the features for retrieval. We randomly choose the average 1000 retrieval times, which takes about 12 ms to compute their similarity. Moreover, since our model learns an excellent intra-class compact and inter-class separable feature distribution, it is very suitable for cross-domain image application scenarios where candidate gallery images can be processed offline, and user query images can be answered on the fly.

## 4.6 Feature embedding using t-SNE

To further investigate why the MMFL-Net functions effectively, we use t-SNE to visualize the extraction feature embedding in a 2D spaces. In Fig. 12, we randomly select nine PIDs from DeepFashion-C2S embedded using MMFL-Net, GF-branch, PHF-branch, PVF-branch, and LDF-branch. Fig. 12(a) shows the effect of embedding using MMFL-Net. The feature embedding is more clustered with some PIDs, and the embedding is well separated for different PIDs. Fig. 12(b) shows the effect of embedding by the GF-branch, which is lower than that of MMFL-Net, and focus of the GF-branch on global salient semantic information, which is relatively higher. Figs. 12(c) and 12(d) show the effect of embedding via the PHF-branch/PVF-branch, where the part-based component branch localizes to the local region. However, owing to the part-based partition, the receptive field of the feature maps decreases. For example, there is a significant difference between hard-view, hard-occlusion, and hard-cropping in images PID 4 and PID 6, resulting in some outliers that increase the intra-class distance. Fig. 12(e) depicts the embedding generated with the LDF-branch. Because the branch focus on local details and lacks a global perspective, its feature embedding is more compact than that of other branches; however, the separability between the classes is not strong. For example, for PID 0 and PID 7, the intra-class distances are minor when compared to those of MMFL-Net. The local detailed fashion similarity is complementary to the global fashion similarity.

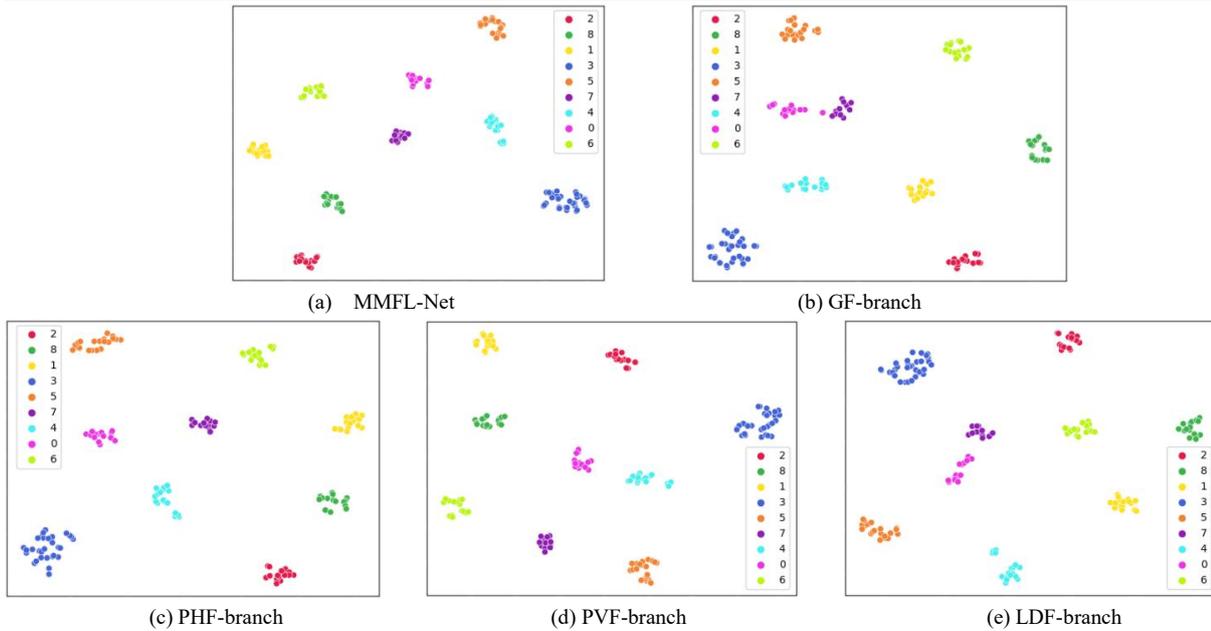

Fig. 12 t-SNE based visualization of random sample embedding space generated using MMFL-Net, GF-branch, PHF-branch, PVF-branch, and LDF-branch; dots with the same color indicate images in the same PIDs.

## 5 Conclusions

In this paper, we explore the practical scenario task of consumer-to-shop visual search and propose a new multi-scale and multi-granularity feature learning network architecture (MMFL-Net) as an end-to-end cross-domain visual similarity model for extracting a more discriminative and robust feature representations. We first design an SFFP structure to apply top-down and bottom-up bidirectional feature fusion scheme to extract multi-scale visual features with global context information. Then, we propose a multi-branch deep network architecture to capture global-local aggregation features for multi-task feature learning. Each branch in the MMFL-Net learns diverse feature representations with multi-granularity. Meanwhile, we jointly optimize the distribution of sample features in the embedding space by combining multi-task classification loss and an improved hard-mining trihard loss and center loss. We use the experimental results demonstrate that the proposed model improves retrieval performance of the consumer-to-shop clothing retrieval task.

In future work, effectiveness and efficient feature fusion approaches and better ways to obtain multi-granularity features or the use of pure transformer-based architecture to solve the problems of clothing deformation and occlusion remain to be studied. For the clothing attribute recognition and retrieval system, some novel multi-task classification and metric learning scheme for feature learning can be introduced to extract more robust and discriminative feature representations so as to facilitate the online clothing shopping.

## Acknowledgements

The authors would like to thank the anonymous reviewers for their helpful and valuable comments and suggestions. This work was supported in part by the National Natural Science Foundation of China under Grant No. 61972458, 62172367, in part by the Zhejiang Province Public Welfare Technology Application Research Project under Grant LGF22F020006.

# Declarations

**Conflicts of interest** the authors declare that they have no conflict of interest.

# Data Availability

The datasets generated during the current study are available from the corresponding author on reasonable request.

# References

1. Cheng WH, Song S, Chen CY, Hidayati SC, Liu J (2021) Fashion meets computer vision: A survey. ACM Computing Surveys, 54(4):1–41
2. Yang F, Kale A, Bubnov Y, Stein L, Wang Q, Kiapour H, Piramuthu R (2017) Visual search at ebay. In: Proceedings of the 23rd ACM SIGKDD international conference on knowledge discovery and data mining, pp 2101–2110
3. Chen W, Liu Y, Wang W, Bakker E, Georgiou T, Fieguth P, Liu L, Lew MS (2021) Deep learning for instance retrieval: A survey. arXiv preprint arXiv: 2101.11282
4. Di W, Wah C, Bhardwaj A, Piramuthu R, Sundaresan N (2013) Style finder: Fine-grained clothing style recognition and retrieval. In: IEEE conference on computer vision and pattern recognition workshops, pp 8–13
5. Bossard L, Dantone M, Leistner C, Wengert C, Quack T, Gool LV (2012) Apparel classification with style. In: Asian conference on computer vision, pp 321–335
6. Kiapour MH, Han X, Lazebnik S, Berg AC, Berg TL (2015) Where to buy it: Matching street clothing photos in online shops. In: Proceedings of the IEEE international conference on computer vision, pp 3343–3351
7. Huang J, Feris RS, Chen Q, Yan S (2015) Cross-domain image retrieval with a dual attribute-aware ranking network. In: Proceedings of the IEEE international conference on computer vision, pp 1062–1070
8. Liu Z, Luo P, Qiu S, Wang X, Tang X (2016) DeepFashion: Powering robust clothes recognition and retrieval with rich annotations. In: Proceedings of the IEEE conference on computer vision and pattern recognition, pp 1096–1104
9. Kuang Z, Gao Y, Li G, Luo P, Chen Y, Lin L, Zhang W (2019) Fashion retrieval via graph reasoning networks on a similarity pyramid. In: Proceedings of the IEEE/CVF International Conference on Computer Vision, pp 3066–3075
10. Hidayati SC, You CW, Cheng WH, Hua KL (2018) Learning and recognition of clothing genres from full-body images. IEEE transaction on cybernetics, 48(5):1647–1659
11. Wieczorek M, Michalowski A, Wroblewska A, Dabrowski J (2020) A strong baseline for fashion retrieval with person re-Identification models. In: International Conference on Neural Information Processing, pp 294–301
12. Han X, Wu Z, Jiang YG, Davis LS (2017) Learning fashion compatibility with bidirectional lstms. In: Proceedings of the 25th ACM International conference on Multimedia, pp 1078–1086
13. Lu S, Zhu X, Wu Y, Wan X, Gao F (2021) Outfit compatibility prediction with multi-layered feature fusion network. Pattern Recognition Letters 147:150–156
14. Zhang X, Chen Y, Zhu B, Wang J, Tang M (2020) Semantic-spatial fusion network for human parsing. Neurocomputing, 402:375–383
15. Liu S, Song Z, Liu G, Xu C, Lu H, Yan S (2012) Street-to-shop: Cross-scenario clothing retrieval via parts alignment and auxiliary set. In: Proceedings of the IEEE conference on computer vision and pattern recognition, pp 3330–3337
16. Biasotti S, Cerri A, Aono M et al (2016) Retrieval and classification methods for textured 3D models: a comparative study. The Visual Computer, 32(2):217–241
17. Biasotti S, Cerri A, Abdelrahman M et al (2014) SHREC'14 track: Retrieval and classification on textured 3D models. In: Proceedings of the Eurographics workshop on 3d object retrieval, pp 111–120
18. Kuo YH, Hsu WH (2017) Feature learning with rank-based candidate selection for product search. In: Proceedings of the IEEE International Conference on Computer Vision Workshops, pp 298–307
19. Kucer M, Murray N (2019) A detect-then-retrieve model for multi-domain fashion item retrieval. In: Proceedings of the IEEE/CVF Conference on Computer Vision and Pattern Recognition Workshops, pp 344–353
20. Lang Y, He Y, Yang F, Dong J, Xue H (2020) Which is plagiarism: Fashion image retrieval based on regional representation for design protection. In: Proceedings of the IEEE/CVF Conference on Computer Vision and Pattern Recognition, pp 2595–2604
21. Liu AA, Zhang T, Song D, Li W, Zhou M (2021) FRSFN: A semantic fusion network for practical fashion retrieval. Multimedia Tools and Applications, 80(11):17169–17181
22. Sun Y, Zheng L, Yang Y, Tian Q, Wang S (2018) Beyond part models: Person retrieval with refined part pooling (and a strong convolutional baseline). In: Proceedings of the European Conference on computer vision, pp 480–496
23. Wang G, Yuan Y, Chen X, Li J, Zhou X (2018) Learning discriminative features with multiple granularities for person re-identification. In: Proceedings of the 26th ACM international conference on Multimedia, pp 274–282
24. Luo Z, Yuan J, Yang J, Wen W (2019) Spatial constraint multiple granularity attention network for clothes retrieval. In: 26th IEEE International Conference on Image Processing, pp 859–863
25. Chen Q, Huang J, Feris R, Brown LW, Dong J, Yan S (2015) Deep domain adaptation for describing people based on fine-grained clothing attributes. In: Proceedings of the IEEE conference on computer vision and pattern recognition, pp 5315–5324
26. Ak KE, Kassim AA, Lim JH, Tham JY (2018) Learning attribute representations with localization for flexible fashion search. In: Proceedings of the IEEE conference on computer vision and pattern recognition, pp 7708–7717
27. Hou Y, Vig E, Donoser M, Bazzani, L (2021) Learning attribute-driven disentangled representations for interactive fashion retrieval. In: Proceedings of the IEEE/CVF International Conference on Computer Vision, pp 12147–12157
28. Ji X, Wang W, Zhang M, Yang Y (2017) Cross-domain image retrieval with attention modeling. In: Proceeding of the 25th ACM international conference on Multimedia, pp 1654–1662
29. Luo Y, Wang Z, Huang Z, Yang Y, Lu H (2019) Snap and find: Deep discrete cross-domain garment image retrieval. arXiv preprint arXiv:1904.02887
30. Wang Z, Gu Y, Zhang Y, Zhou J, Gu X (2017) Clothing retrieval with visual attention model. In: 2017 IEEE Visual Communications and Image Proceeding, pp 1–4
31. Dong J, Ma Z, Mao X, Yang X, He Y, Hong R, Ji S (2021) Fine-grained fashion similarity prediction by attribute-specific embedding learning. IEEE Transaction on Image Processing, 30:8410–8425
32. Tan F, Yuan J, Ordonez V (2021). Instance-level image retrieval using reranking transformers. In: Proceedings of the IEEE/CVF International Conference on Computer Vision, pp 12105–12115

<em>Multimedia Tools and Applications (2022)</em>

33. Newell A, Yang K, Deng J (2016) Stacked hourglass networks for human pose estimation. In: European conference on computer vision, pp 483–499
34. Lin TY, Dollar P, Girshick R, He K, Hariharan B, Belongie S (2017) Feature pyramid networks for object detection. In: Proceedings of the IEEE conference on computer vision and pattern recognition, pp 936–944
35. Liu S, Qi L, Qin H, Shi J, Jia J (2018) Path aggregation network for instance segmentation. In: Proceedings of the IEEE conference on computer vision and pattern recognition, pp 8759–8768
36. Tan M, Pang R, Le QV (2020) Efficientdet: Scalable and efficient object detection. In: Proceedings of the IEEE/CVF conference on computer vision and pattern recognition, pp 10781–10790
37. Zhao Q, Sheng T, Wang Y, Tang Z, Chen Y, Cai L, Ling H (2019) M2det: A single-shot object detector based on multi-level feature pyramid network. In: Proceedings of the AAAI conference on artificial intelligence, 33(1), pp 9259–9266
38. Zhang Z, Zhang X, Peng C, Xue X, Sun J (2018) Exfuse: Enhancing feature fusion for semantic segmentation. In: Proceedings of the European conference on computer Vision, pp 269–284
39. Wang Q, Wu B, Zhu P, Li P, Zuo W, Hu Q (2020) ECA-Net: Efficient channel attention for deep convolutional neural networks. In: Proceedings of the IEEE/CVF Conference on Computer Vision and Pattern Recognition, pp 11531–11539
40. Wang F, Jiang M, Qian C, Yang S, Li C, Zhang H, Wang X, Tang X (2017) Residual attention network for image classification. In: Proceedings of the IEEE conference on computer vision and pattern recognition, pp 3156–3164
41. Yang M, Yu K, Zhang C, Li Z, Yang K (2018) Denseaspp for semantic segmentation in street scenes. In: Proceedings of the IEEE conference on computer vision and pattern recognition, pp 3684–3692
42. Zhao H, Shi J, Qi X, Wang X, Jia J (2017) Pyramid scene parsing network. In: Proceedings of the IEEE conference on computer vision and pattern recognition, pp 2881–2890
43. Cao Y, Xu J, Lin S, Wei F, Hu H (2019) Gcnet: Non-local networks meet squeeze-excitation networks and beyond. In: Proceedings of the IEEE/CVF international conference on computer vision workshops, pp 1971–1980
44. Lu Y, Kumar A, Zhai S, Cheng Y, Javidi T, Feris R (2017) Fully-adaptive feature sharing in multi-task networks with applications in person attribute classification. In: Proceedings of the IEEE conference on computer vision and pattern recognition, pp 5334–5343
45. Sharma V, Murray N, Larlus D et al (2021) Unsupervised meta-domain adaptation for fashion retrieval. In: Proceedings of the IEEE/CVF Winter Conference on Applications of Computer Vision, pp 1348–1357
46. Zhang H, Cisse M, Dauphin YN, Lopez-Paz D (2017) mixup: Beyond empirical risk minimization. arXiv preprint arXiv:1710.09412
47. Zhong Z, Zheng L, Cao D, Li S (2017) Re-ranking person re-identification with k-reciprocal encoding. In: Proceedings of the IEEE conference on computer vision and pattern recognition, pp 1318–1327
48. Veit A, Belongie S, Karaletsos T (2017) Conditional similarity networks. In: Proceedings of the IEEE conference on computer vision and pattern recognition, pp 830–838
49. Wieczorek M, Rychalska B, Dabrowski J (2021) On the unreasonable effectiveness of centroids in image retrieval. In: International Conference on Neural Information Processing, pp 212–223
50. Dong Q, Gong S, Zhu X (2017) Multi-task curriculum transfer deep learning of clothing attributes. In: 2017 IEEE Winter Conference on Applications of Computer Vision, pp 520–529
51. Wang Z, Pu Y, Wang X, Zhao Z, Xu D, Qian W (2020) Accurate Retrieval of Multi-scale clothing images based on multi-feature fusion. Chinese Journal of Computers, 43(4):740–754
52. Selvaraju RR, Cogswell M, Das A, Vedantam R, Parikh D, Batra D (2017) Grad-cam: Visual explanations from deep networks via gradient-based localization. In: Proceedings of the IEEE international conference on computer vision, pp 618–626